\documentclass[10pt,twocolumn,letterpaper]{article}

\usepackage{cvpr}              % To produce the CAMERA-READY version
\usepackage[accsupp]{axessibility}
\usepackage{pifont} % For tick and cross symbols
\newcommand{\xmark}{\ding{55}}  % Cross symbol
\usepackage{makecell}

\usepackage{graphicx}
\usepackage{colortbl}
\usepackage{amsmath}
\usepackage{amssymb}
\usepackage{booktabs}

\definecolor{cvprblue}{rgb}{0.21,0.49,0.74}
\usepackage[pagebackref,breaklinks,colorlinks,allcolors=cvprblue]{hyperref}

\def\paperID{*****} % *** Enter the Paper ID here
\def\confName{CVPR}
\def\confYear{2025}

\title{FUSION: Frequency-guided Underwater Spatial Image recOnstructioN}

\author{
    Jaskaran Singh Walia\thanks{Equal contribution} \quad 
    Shravan Venkatraman\footnotemark[1] \quad 
    Pavithra L K\\
    Vellore Institute of Technology, Chennai, India \\
    {\tt\small corresponding email: pavithra.lk@vit.ac.in} \\
    {\tt\small * Equal contribution} \\
}

\begin{document}

% Use \twocolumn[{...}] to ensure the figure appears right after the title
\twocolumn[{%
    \renewcommand\twocolumn[1][]{#1}%
    \maketitle
    \begin{center}
        \captionsetup{type=figure}  % Ensure figure numbering is correct
        \includegraphics[width=\linewidth]{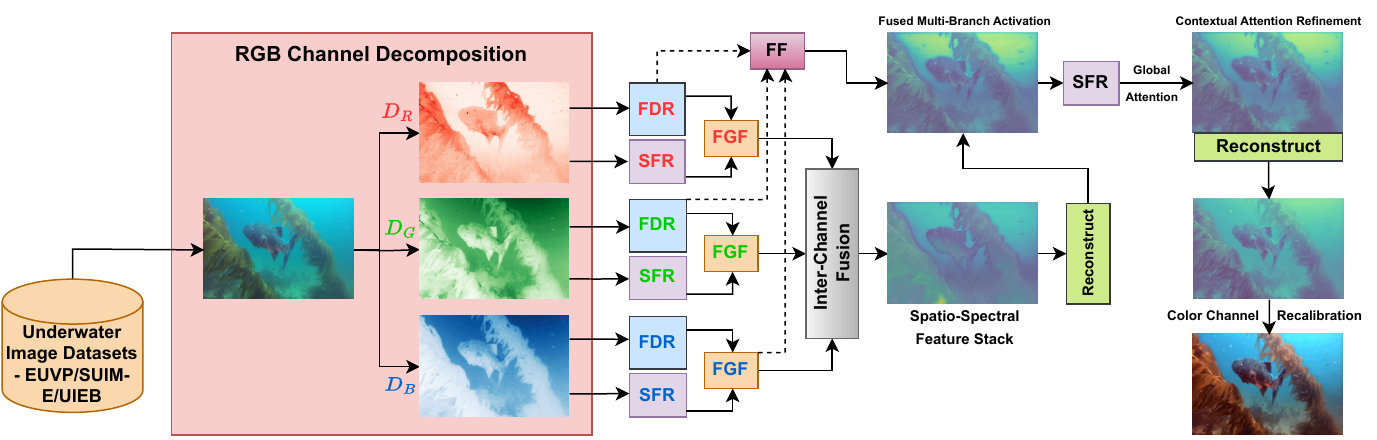}
        \caption{An overview of the proposed FUSION pipeline, illustrating the dual-domain (spatial and frequency) processing, contextual attention refinement, and final channel calibration for UIE.}
        \label{fig:pipeline}
    \end{center}
}]

\begin{abstract}
Underwater images suffer from severe degradations, including color distortions, reduced visibility, and loss of structural details due to wavelength-dependent attenuation and scattering. Existing enhancement methods primarily focus on spatial-domain processing, neglecting the frequency domain's potential to capture global color distributions and long-range dependencies. To address these limitations, we propose FUSION, a dual-domain deep learning framework that jointly leverages spatial and frequency domain information. FUSION independently processes each RGB channel through multi-scale convolutional kernels and adaptive attention mechanisms in the spatial domain, while simultaneously extracting global structural information via FFT-based frequency attention. A Frequency Guided Fusion module integrates complementary features from both domains, followed by inter-channel fusion and adaptive channel recalibration to ensure balanced color distributions. Extensive experiments on benchmark datasets (UIEB, EUVP, SUIM-E) demonstrate that FUSION achieves state-of-the-art performance, consistently outperforming existing methods in reconstruction fidelity (highest PSNR of 23.717 dB and SSIM of 0.883 on UIEB), perceptual quality (lowest LPIPS of 0.112 on UIEB), and visual enhancement metrics (best UIQM of 3.414 on UIEB), while requiring significantly fewer parameters (0.28M) and lower computational complexity, demonstrating its suitability for real-time underwater imaging applications.
  % Abstract follows immediately after the figure
\end{abstract}

\section{Introduction}
\label{sec:intro}
Underwater imaging plays a crucial role in fields like marine biology, underwater archaeology, and autonomous underwater vehicle (AUV) navigation. However, it faces challenges such as light absorption and scattering, leading to low contrast, color casts (bluish and greenish hues), and blurriness, which hinder high-level vision tasks like object detection and segmentation \cite{1,2,3}. Traditional underwater image enhancement (UIE) methods, such as histogram equalization and dehazing algorithms, struggle with the complex degradations in underwater environments \cite{4}. Advanced cameras also fail to address non-uniform light attenuation, where shorter wavelengths like blue and green penetrate deeper underwater, distorting color balance and reducing task performance \cite{5}.

Deep learning-based techniques have recently shown promise in low-level vision tasks. State-of-the-art underwater image restoration (UIR) methods often use identical receptive field sizes for R-G-B channels, ignoring wavelength-dependent degradation patterns. Sharma et al. \cite{6} demonstrated that varying receptive field sizes (e.g., R$(3\times3)$, G$(5\times5)$, B$(7\times7)$) improves UIR by capturing local and global features. Encoder-decoder networks commonly used in UIR capture broad contexts but lose spatial details during downsampling \cite{7,8}. High-resolution networks avoid downsampling but struggle with encoding global context needed for coherent enhancement. Most UIR methods focus solely on spatial-domain processing, overlooking long-range dependencies and global color distributions essential for effective UIE.

To address these limitations, we propose FUSION: Frequency-guided Underwater Spatial Image recONstructioN—a dual-domain framework tailored for underwater image enhancement. FUSION integrates spatial and frequency domain processing through four key modules: the Multi-Scale Spatial Module processes RGB channels using dedicated kernel sizes to handle wavelength-dependent attenuation; the Frequency Extraction Module refines magnitude information to capture global structural cues; the Frequency-Guided Fusion (FGF) Module combines spatial and frequency features for balanced local detail and global color consistency; and the Inter-Channel Fusion and Channel Calibration Module uses global attention and adaptive scaling to produce enhanced images with balanced color distribution.

Our dual-path architecture (Figure~\ref{fig:pipeline}) processes RGB channels ($D_R$, $D_G$, $D_B$) independently. Spatial features are refined using multi-scale convolutions and attention mechanisms, while frequency features are extracted using Fourier analysis to capture global information. Spatial-frequency features are fused via FGF blocks for each channel before inter-channel fusion integrates RGB dependencies. A decoder stage with deconvolutional layers, attention mechanisms, residual connections, and adaptive recalibration balances RGB channels to produce enhanced images with improved visibility, color accuracy, and detail preservation.

To summarize, our contributions are as follows:
\begin{itemize}
    \item \textbf{Dual-Domain Enhancement:} We introduce a parallel frequency pathway that captures long-range dependencies and global color distributions, complementing traditional spatial processing.
    \item \textbf{Dedicated Frequency Attention Module:} By preserving original phase while applying adaptive attention to the magnitude spectrum, our method captures global structural information critical for handling complex underwater degradations.
    \item \textbf{Inter-Channel Calibration for Color Correction:} A global recalibration stage, which employs learnable scaling factors to balance color intensities adaptively. 
\end{itemize}

\section{Related Works}
\label{sec:relatedWorks}

\subsection{Underwater Image Enhancement}

Traditional methods for UIE have relied on image processing techniques such as histogram equalization, white balance adjustment, and dehazing algorithms based on physical models of light propagation underwater \cite{turn0search0, turn0search6, turn0search5}. While these methods can enhance contrast and correct color casts to some extent, they generally lack adaptability to varying underwater conditions and often fail to restore fine details and textures. Li et al. proposed a dehazing and color correction method using convolutional neural networks (CNNs) that leverage the statistical properties of underwater images \cite{turn0search5}. FUnIE-GAN and Water-Net have shown promising results by learning mappings from degraded images to their enhanced counterparts using generative adversarial networks (GANs) \cite{turn0search7, turn0search8}. 

There exists minimal literature on frequency-based methods for image enhancement, with no prior application to underwater imaging. Kersting et al. used a GAN-based approach to enhance ultra-fast PSMA-PET scans via synthetic reconstruction, showing improved detection in prostate cancer staging \cite{freqlit1}. Liu et al. applied frequency decomposition in PID$^2$Net for underwater descattering and denoising, though not explicitly using frequency-domain learning \cite{freqlit2}. Li et al. fused polarization cues with wavelet-based subband processing to improve defect visibility on reflective surfaces \cite{freqlit3}. Agaian et al. proposed transform-based enhancement using orthogonal bases like Fourier and Hadamard with quantifiable performance metrics \cite{freqlit4}. Wang proposed a parallel frequency-domain low-light framework that decouples contrast and structure restoration \cite{freqlit5}, while Wang et al. designed FourLLIE, leveraging Fourier amplitude mappings and SNR-guided fusion for efficient low-light enhancement \cite{freqlit6}.

\subsection{Attention Mechanisms in Image Enhancement}

Attention mechanisms are integrated into deep learning models to improve feature representation by focusing on the most informative parts of the input. In the context of image enhancement, attention modules can help models learn where to emphasize or suppress features, leading to better restoration of degraded images.

Chen et al. introduced an attention-based UIE method that employs a multi-scale attention mechanism to adaptively enhance features at different resolutions \cite{turn0search7, turn0search8}. Similarly, Li et al. utilized channel attention in their network to weigh the importance of different feature maps, improving the overall enhancement quality. While these methods have shown effectiveness, they often increase the model's complexity and computational load \cite{turn0search1, turn0search5}.

\subsection{Shortcomings Addressed}

The primary limitation in current underwater image restoration and enhancement approaches is that they focus predominantly on spatial-domain processing, overlooking the frequency domain’s ability to capture global color distributions and long-range dependencies. This omission often results in residual color imbalances and artifacts, especially under severe wavelength-dependent attenuation~\cite{turn0search1}. Additionally, certain models that are able to capture these domains (like Fine-tuned GANs) require very heavy computational power, which makes them not viable for deployment and scalability scenarios~\cite{turn0search5}. Our proposed FUSION addresses these issues through a dual-domain design that efficiently processes each color channel in both spatial and frequency domains while having a quick inference time and low-memory compute. By incorporating multi-stage and multi-domain attention mechanisms with channel-wise recalibration, FUSION also preserves fine details, reduces artifacts, and balances color distributions.

\section{Proposed Method: \textit{FUSION}}
\label{sec:proposedMethod}

\begin{figure*}[t]
  \centering
   \includegraphics[width=\linewidth]{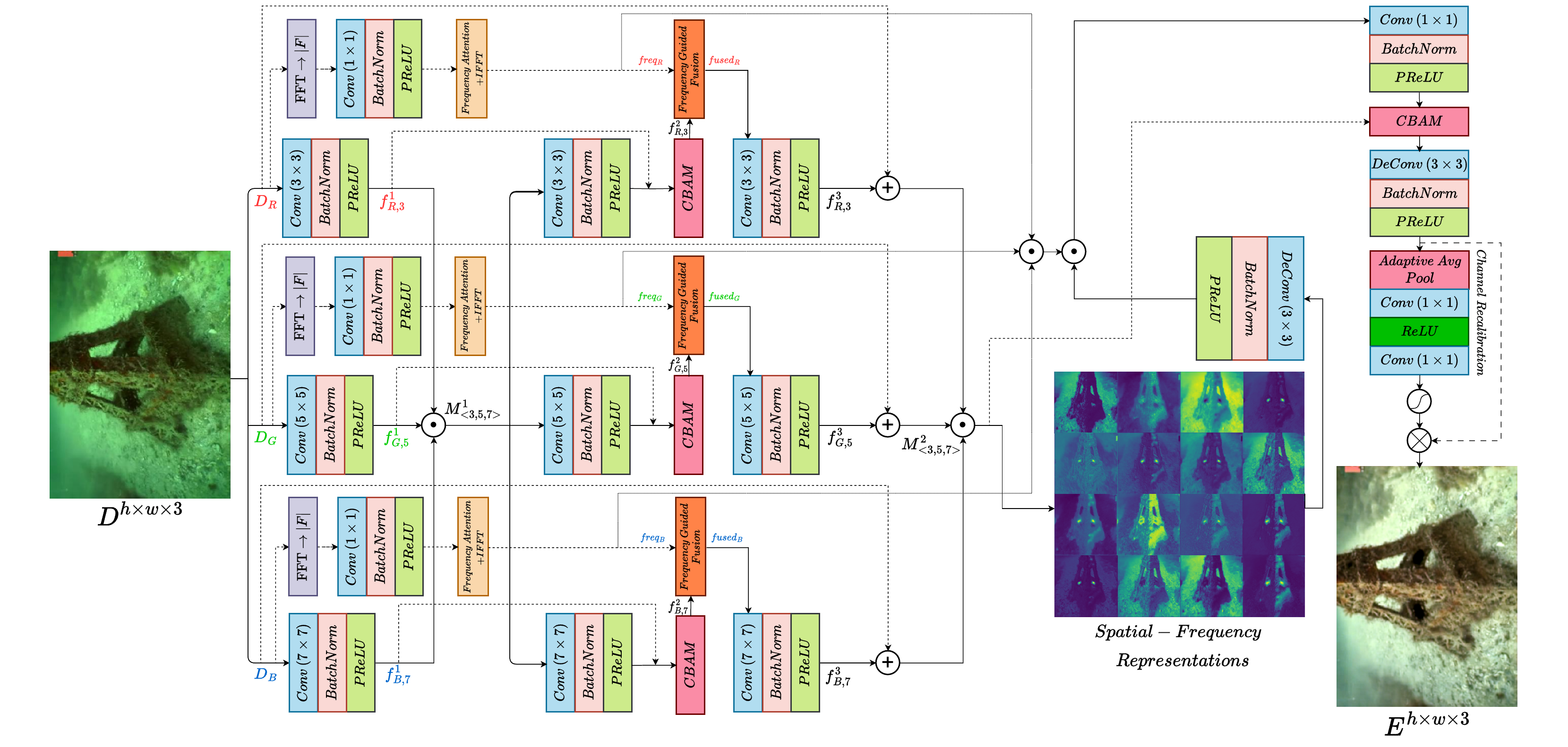}
   \caption{Overview of our proposed FUSION architecture for UIE. The model takes a degraded underwater image as input and restores it with enhanced visual quality.}
   \label{fig:architecture}
\end{figure*}

The proposed architecture enhances underwater images through a dual-path framework that integrates spatial domain processing and frequency domain processing. The input image \( D^{h \times w \times 3} \) is split into three independent channels, \( D_R \), \( D_G \), and \( D_B \), which are processed separately in both domains to extract complementary features. 

In the spatial domain, each channel undergoes multi-scale convolution with kernel sizes \( 3 \times 3 \), \( 5 \times 5 \), and \( 7 \times 7 \) to capture features at varying receptive fields. This ensures that color-specific distortions are addressed independently, preventing the propagation of noisy features while preserving crucial wavelength-driven contextual information, as suggested in \cite{sharma2023wavelength}. These features are refined using a Channel and Spatial Attention Module (CBAM) and residual connections to preserve information.

In parallel, frequency domain features are extracted by transforming each channel into the frequency domain using a 2D Fast Fourier Transform (FFT). The magnitude of the frequency representation is processed using $1 \times 1$ convolutional layers and refined with a Frequency Attention mechanism. The inverse FFT (IFFT) reconstructs these features back into the spatial domain.

The outputs from the spatial (\( f^3_{R/G/B,\dots} \)) and frequency (\( freq_{R/G/B,\dots} \)) domains are fused using FGF blocks. Finally, the fused features are passed through a decoder with a global attention (CBAM) and channel recalibration to adaptively balance RGB channels, producing the enhanced underwater image $E^{h \times w \times 3}$.

\subsection{Spatial Domain Processing}
The spatial domain processing path extracts features from each channel of the input image $D^{h \times w \times 3}$ by leveraging multi-scale feature extraction, attention mechanisms, and residual refinement. Each channel, $D_R$, $D_G$, and $D_B$, is processed independently to capture spatial patterns at multiple scales $\{s_1, s_2, s_3\}$.

Initially, feature maps $f^1_{i} = \Phi_i(D_i)$ are extracted from each channel $i \in \{R,G,B\}$ using convolutional operations with varying receptive fields. Specifically, $f^1_{R}$ represents the features extracted from the red channel using kernel size $3 \times 3$, while $f^1_{G}$ and $f^1_{B}$ are obtained with $5 \times 5$ and $7 \times 7$ kernels, respectively. This multi-scale extraction $\{f^1_R, f^1_G, f^1_B\}$ enables the network to capture hierarchical features across the feature dimension with varying spatial contexts.

To enhance these features, a two-stage attention mechanism $\mathcal{A} = \mathcal{A}_c \circ \mathcal{A}_s$ is applied independently to each channel. In the first stage, channel attention $\mathcal{A}_c$ aggregates global information by computing scaling weights $W_{\text{channel}}$ based on pooled statistics of the feature map:
\begin{equation}
    W_{\text{channel}} = \sigma\big(\mathbf{W}_2 \cdot \phi(\mathbf{W}_1 \cdot g(f^1_i))\big)
\end{equation}
Here $g(f^1_i)$ represents global average pooling, $\phi(\cdot)$ implements ReLU activation, and $\mathbf{W}_1$, $\mathbf{W}_2$ are learnable weight matrices with reduction ratio $r$. The feature map is then scaled element-wise as $f_{\text{channel-att}} = W_{\text{channel}} \odot f^1_i$.

In the second stage, spatial attention $\mathcal{A}_s$ refines these channel-weighted features by focusing on spatially significant regions through attention mapping. This is achieved by computing spatial attention weights:
\begin{equation}
    W_{\text{spatial}} = \sigma\big(h(f_{\text{channel-att}})\big)
\end{equation}
\begin{equation}
    h(f_{\text{channel-att}}) = 
    \psi\Big( [
    \mathcal{P}_{avg}(f_{\text{channel-att}}); 
    \mathcal{P}_{max}(f_{\text{channel-att}}) ] \Big)
\end{equation}
The function $h$ aggregates information across channels via concatenated max and average pooling operations, followed by a spatial transformation. The final attention-refined feature map is given by $f_{\text{spatial-att}} = W_{\text{spatial}} \odot f_{\text{channel-att}}$.

After applying both attention mechanisms, the refined feature maps for each channel are denoted as $f^2_{i} = \mathcal{A}(f^1_i) = \mathcal{A}_s(\mathcal{A}_c(f^1_i))$ for $i \in \{R,G,B\}$. To preserve original spatial information and improve gradient flow during training, residual connections are added:
\begin{equation}
    f^3_{i} = f^2_{i} + f^1_{i} \quad \forall i \in \{R,G,B\}
\end{equation}
These skip connections ensure that low-level features are preserved throughout the network while allowing the learning of residual mappings. The outputs, $f^3_{R}$, $f^3_{G}$, $f^3_{B}$, represent the final spatial representations for each channel after multi-scale feature extraction, attention-based refinement, and residual enhancement.

By processing each color channel independently, we address the unique degradation patterns in underwater images where different wavelengths of light are attenuated at rates dependent on depth and water properties. The multi-scale feature extraction with varying kernel sizes is specifically designed to capture the diverse spatial characteristics present in underwater scenes, from fine-grained textures to broader structural elements.

\subsection{Frequency Domain Processing}
The frequency domain processing path complements the spatial domain by extracting and refining frequency features from each channel of the input image $D^{h \times w \times 3}$. This path leverages Fourier transforms, magnitude extraction, frequency attention, and inverse reconstruction to capture global contextual information that is often inaccessible in the spatial domain.

Each channel, $D_i$ for $i \in \{R,G,B\}$, is independently transformed into the frequency domain using a 2D Fast Fourier Transform (FFT). For a given channel, the FFT produces a complex-valued representation $F_i = \mathcal{F}(D_i)$ containing both real and imaginary components. The magnitude of this representation is extracted as:
\begin{equation}
|F_i| = \sqrt{\text{Re}(F_i)^2 + \text{Im}(F_i)^2}
\end{equation}
This magnitude $|F_i|$ captures global structural information about the input channel, where $\text{Re}(F_i)$ and $\text{Im}(F_i)$ denote the real and imaginary parts of the frequency representation. To refine these magnitude features, we apply a series of transformations in the frequency domain. The magnitude map $|F_i|$ undergoes linear transformations with learnable weight matrices $W_1$ and $W_2$ to reduce dimensionality and enhance discriminative features:
\begin{equation}
\hat{F}_i = W_2 \cdot \phi(W_1 \cdot |F_i|)
\end{equation}

\begin{figure}[t]
  \centering
   \includegraphics[width=\linewidth]{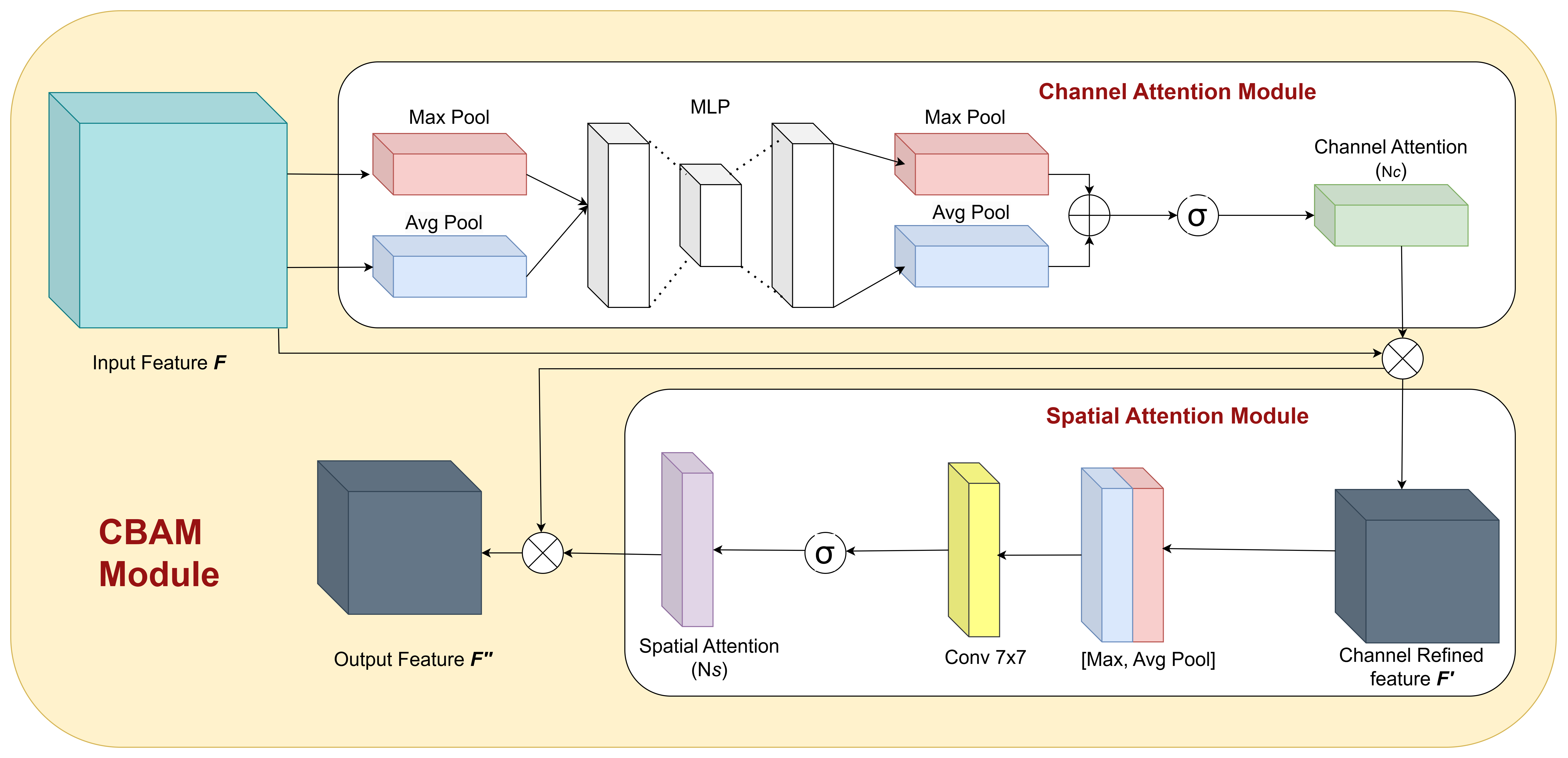}
   \caption{Architecture of the CBAM block \cite{cbam}}
   \label{fig:cbam}
\end{figure}

These transformations incorporate PReLU activation function $\phi(\cdot)$ and are followed by normalization to stabilize feature distributions across varying underwater conditions. Since underwater images suffer from wavelength-dependent attenuation that manifests differently in the frequency spectrum, these transformations help isolate discriminative frequency features that carry reliable information about the scene. A Frequency Attention Module further enhances these features by computing attention weights $W_{\text{freq}}$ for each channel:
\begin{equation}
W_{\text{freq}} = \sigma(W_4 \cdot \phi(W_3 \cdot g(|F_i|)))
\end{equation}
Here $g(|F_i|)$ represents global average pooling, $W_3$ and $W_4$ are learnable weights, and $\sigma(\cdot)$ denotes sigmoid activation. The refined magnitude map $|F_i|_{\text{refined}} = W_{\text{freq}} \odot |F_i|$ adaptively amplifies important frequency components while suppressing less informative ones. This attention mechanism is particularly crucial for underwater imagery where certain frequency bands may be more degraded than others depending on water properties and depth. The refined magnitude is then recombined with the original phase information $\Theta_i = \text{Phase}(F_i)$ to reconstruct a complex-valued frequency representation:
\begin{equation}
F'_i = |F_i|_{\text{refined}} \cdot e^{j\cdot\Theta_i}
\end{equation}
This phase preservation is essential as it maintains structural coherence while allowing magnitude enhancement. The exponential phase term can be expressed as $e^{j\cdot\Theta_i} = \cos(\Theta_i) + j\cdot\sin(\Theta_i)$ where $\Theta_i = \arctan\left(\frac{\text{Im}(F_i)}{\text{Re}(F_i)}\right)$. Finally, an inverse FFT (IFFT) transforms the refined frequency representation back into the spatial domain:
\begin{equation}
f_{\text{freq},i} = \mathcal{F}^{-1}(F'_i)
\end{equation}
The resulting frequency-derived feature maps, $f_{\text{freq},i}$ for $i \in \{R,G,B\}$, capture global contextual information that complements the localized details extracted in the spatial domain. These frequency features effectively represent long-range dependencies between pixels and global color distributions, which are particularly valuable for underwater image enhancement where visibility degradation affects the entire image non-uniformly.

By processing frequency information independently for each color channel, our approach addresses the channel-specific degradation patterns common in underwater environments, where red wavelengths attenuate more rapidly than green and blue wavelengths with increasing depth according to $I(\lambda,d) = I_0(\lambda)e^{-\beta(\lambda)d}$ \cite{redWavelength}.

\subsection{Frequency Guided Fusion}
We integrate spatial and frequency features through our FGF blocks, which operate independently for each channel (Red, Green, Blue). These blocks combine complementary information from spatial domain ($f_{\text{spatial},i}$) and frequency domain ($f_{\text{freq},i}$) to produce fused features $f_{\text{fused},i}$ for each channel $i \in \{R, G, B\}$.

For each color channel, we first concatenate the spatial feature map $f_{\text{spatial},i}$ and the frequency feature map $f_{\text{freq},i}$ along the channel dimension:
\begin{equation}
f_{\text{concat},i} = \mathcal{C}(f_{\text{spatial},i}, f_{\text{freq},i})
\end{equation}
This creates a unified representation containing both local spatial details and global frequency characteristics crucial for underwater image enhancement. We then transform the concatenated feature map through a convolution operation:
\begin{equation}
f_{\text{fused},i} = W_i * f_{\text{concat},i}
\end{equation}
with learnable weights $W_i$ to reduce dimensionality while integrating the two complementary modalities. This ensures that we preserve discriminative features from both domains while managing computational complexity.

The outputs of our FGF blocks, $f_{\text{fused},i}$ for $i \in \{R,G,B\}$, represent channel-specific fused representations that combine both fine-grained spatial details and comprehensive frequency information, capturing both local textures and global color distributions.

\subsection{Inter-Channel Fusion and Channel Calibration}
In the final stage of our architecture, we refine the fused feature representations from each RGB channel to produce the enhanced underwater image $E$. To ensure consistency in feature representation while mitigating underwater distortions, we integrate residual enhancements, spatial-frequency fusion, and adaptive recalibration.

First, we reinforce each fused feature map by adding back the corresponding input channel, ensuring that the residual information is preserved without disrupting learned features:
\begin{equation}
f_{\text{residual},i} = f_{\text{fused},i} + f_{\text{input},i}, \quad i \in \{R, G, B\}
\end{equation}

We concatenate these residual-enhanced features to form a unified representation $f_{\text{concat}}$, enabling our model to leverage inter-channel dependencies effectively. To increase feature expressivity and capture richer spatial characteristics, we project this representation into a higher-dimensional feature space using transformation $\mathcal{T}_d$, yielding:
\begin{equation}
f_d = \phi(\mathcal{T}_d(f_{\text{concat}}))
\end{equation}
where $\phi$ denotes a non-linear activation function. Parallel to this, we extract frequency domain features $f_{\text{freq},i}$ for each RGB channel to capture structural variations that may be less evident in the spatial domain. These features are concatenated as $f_{\text{freq}}$, providing complementary information for the fusion process. To effectively integrate spatial and frequency domain representations, we apply a learned transformation $\mathcal{T}_f$:
\begin{equation}
f_{\text{fusion}} = \phi(\mathcal{T}_f(f_d, f_{\text{freq}}))
\end{equation}

\begin{figure*}[!ht]
    \centering
    \includegraphics[width=\linewidth]{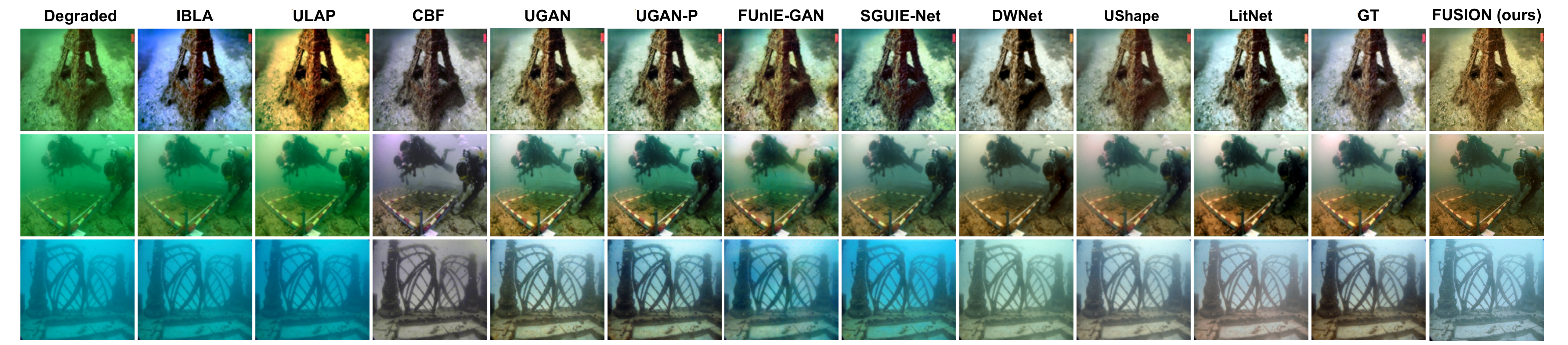}
    \caption{Visual comparisons on the UIEB dataset.}
    \label{fig:visual_UIEB}
\end{figure*}

\begin{figure*}[!h]
    \centering
    \includegraphics[width=\linewidth]{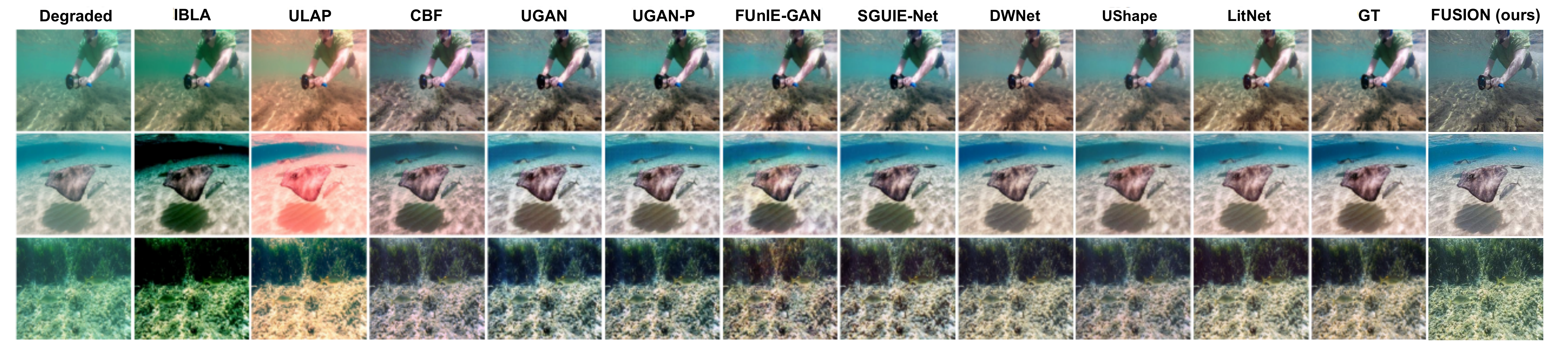}
    \caption{Visual comparisons on the EUVP dataset.}
    \label{fig:visual_EUVP}
\end{figure*}

This allows us to capture localized textures and global structures simultaneously, ensuring effective feature aggregation. Since different regions of the image may require varying levels of enhancement, we refine the fused features using a CBAM-based global attention mechanism $\mathcal{A}$ that selectively emphasizes important regions:
\begin{equation}
f_{\text{attn}} = \mathcal{A}(f_{\text{fusion}}, f_{\text{concat}})
\end{equation}

The attention-refined representation undergoes transformation through $\mathcal{T}_e$, reconstructing a coherent spatial representation $E = \phi(\mathcal{T}_e(f_{\text{attn}}))$. However, this yields pre-channel-calibrated reconstructions, which need further color distribution balancing. To address this and mitigate unwanted shifts, we implement an adaptive recalibration mechanism that generates per-channel scaling factors:
\begin{equation}
W_{\text{calibration}} = \sigma(W_2 \cdot \phi(W_1 \cdot g(E)))
\end{equation}
where $g(E)$ extracts global descriptors summarizing the image's color characteristics, and $\sigma$ normalizes the scaling factors to maintain RGB channel balance. The final enhanced image is obtained through element-wise calibration:
\begin{equation}
E_{\text{final}} = E \odot W_{\text{calibration}}
\end{equation}

This adaptive weighting scheme ensures a visually coherent and perceptually enhanced underwater image by dynamically adjusting color balance and preserving structural details, mitigating common artifacts found in traditional enhancement techniques.

\section{Results}

\paragraph{Experimental Settings}

\begin{table}[ht]
\centering
\caption{Evaluation on UIEB test set with the best-published works for UIE. First, second, and third best performances are represented in \textcolor{red}{red}, \textcolor{blue}{blue}, and \textcolor{green}{green} colors, respectively. ↓ indicates lower is better.}
\resizebox{\columnwidth}{!}{
\begin{tabular}{lcccccc}
\hline
Method & PSNR & SSIM & LPIPS↓ & UIQM & UISM & BRISQUE↓ \\
\hline
UDCP~\cite{drews2013transmission} & 13.026 & 0.545 & 0.283 & 1.922 & 7.424 & 24.133 \\
GBdehaze~\cite{li2016single} & 15.378 & 0.671 & 0.309 & 2.520 & 7.444 & 23.929 \\
IBLA~\cite{peng2017underwater} & 19.316 & 0.690 & 0.233 & 2.108 & 7.427 & 23.710 \\
ULAP~\cite{song2018rapid} & 19.863 & 0.724 & 0.256 & 2.328 & 7.362 & 25.113 \\
CBF~\cite{ancuti2017color} & 20.771 & 0.836 & 0.189 & \textcolor{green}{3.318} & 7.380 & \textcolor{green}{20.534} \\
UGAN~\cite{fabbri2018enhancing} & 23.322 & 0.815 & 0.199 & \textcolor{red}{3.432} & 7.241 & 27.011 \\
UGAN-P~\cite{fabbri2018enhancing} & \textcolor{green}{23.550} & 0.814 & 0.192 & \textcolor{blue}{3.396} & 7.262 & 25.382 \\
FUnIE-GAN~\cite{islam2020fast} & 21.043 & 0.785 & 0.173 & 3.250 & 7.202 & 24.522 \\
SGUIE-Net~\cite{qi2022sguie} & 23.496 & \textcolor{green}{0.853} & \textcolor{green}{0.136} & 3.004 & \textcolor{green}{7.362} & 24.607 \\
DWNet~\cite{sharma2023wavelength} & 23.165 & 0.843 & 0.162 & 2.897 & \textcolor{red}{7.089} & 24.863 \\
Ushape~\cite{peng2023u} & 21.084 & 0.744 & 0.220 & 3.161 & 7.183 & 24.128 \\

Lit-Net \cite{litnet} & \textcolor{blue}{23.603} & \textcolor{blue}{0.863} & \textcolor{blue}{0.130} & 3.145 & \textcolor{blue}{7.396} & \textcolor{red}{23.038} \\

FUSION (Ours) & \textcolor{red}{23.717} & \textcolor{red}{0.883} & \textcolor{red}{0.112} & \textcolor{green}{3.414} & \textcolor{green}{7.429} & \textcolor{blue}{23.193} \\
\hline
\end{tabular}
}
\label{tab:UIEB}
\end{table}

\begin{table*}[!ht]
\centering
\caption{Ablation hardware comparisons with respect to average performance across datasets ($\overline{\mathrm{Metric}}$ denotes the average of that metric across the 3 datasets used).}
\renewcommand{\arraystretch}{1.2}
\resizebox{\textwidth}{!}{%
\begin{tabular}{lccccccccccc}
\hline
\textbf{Configuration} & \textbf{Freq. Attn} & \textbf{Freq. Branch} & \textbf{Freq. Fusion} & \textbf{Chan. Calib} & \textbf{Local Attn} & \textbf{Global Attn} & \textbf{Inference Time (ms)} & GFLOPs & \textbf{$\overline{\mathrm{UISM}}$} & \textbf{$\overline{\mathrm{LPIPS}}$} & $\overline{\mathrm{BRISQUE}}$ \\
\hline
\rowcolor[gray]{0.8}
Full Model (FUSION) & \checkmark & \checkmark & \checkmark & \checkmark & \checkmark & \checkmark & 128.68 & 36.73 & 7.385 & 0.135 & 23.797 \\
No Frequency Attention & \xmark & \checkmark & \checkmark & \checkmark & \checkmark & \checkmark & 128.53 & 36.71 & 6.395 & 0.207 & 27.643 \\
No Frequency Branch & \checkmark & \xmark & \checkmark & \checkmark & \checkmark & \checkmark & 88.89 & 36.55 & 5.996 & 0.255 & 29.553 \\
No Frequency Guided Fusion & \checkmark & \checkmark & \xmark & \checkmark & \checkmark & \checkmark & 90.29 & 36.71 & 6.192 & 0.226 & 28.370 \\
No Channel Calibration & \checkmark & \checkmark & \checkmark & \xmark & \checkmark & \checkmark & 128.70 & 36.73 & 6.164 & 0.230 & 28.517 \\
No Local Attention & \checkmark & \checkmark & \checkmark & \checkmark & \xmark & \checkmark & 75.87 & 36.69 & 6.453 & 0.210 & 27.627 \\
No Global Attention & \checkmark & \checkmark & \checkmark & \checkmark & \checkmark & \xmark & 110.69 & 36.72 & 6.561 & 0.200 & 27.167 \\
Spatial Only & \xmark & \xmark & \xmark & \checkmark & \checkmark & \checkmark & 89.01 & 36.55 & 5.908 & 0.250 & 29.320 \\
Minimal Model & \xmark & \xmark & \xmark & \xmark & \xmark & \xmark & 18.49 & 36.49 & 5.704 & 0.276 & 30.607 \\
\hline
\end{tabular}%
}
\label{tab:ablation_hardware_multiline}
\end{table*}

\begin{table}[ht]
\centering
\caption{Evaluation on EUVP dataset with the best-published works for UIE. First, second, and third best performances are represented in \textcolor{red}{red}, \textcolor{blue}{blue}, and \textcolor{green}{green} colors, respectively. ↓ indicates lower is better.}
\resizebox{\columnwidth}{!}{
\begin{tabular}{lccccccc}
\hline
Method & MSE↓ & PSNR & SSIM & UIQM & LPIPS↓ & UISM & BRISQUE↓ \\
\hline
UGAN~\cite{fabbri2018enhancing}         & 0.355                      & 26.551                     & 0.807                     & 2.896                     & 0.220                     & 6.833                     & 35.859 \\
UGAN-P~\cite{fabbri2018enhancing}       & 0.347                      & 26.549                     & 0.805                     & 2.931                     & 0.223                     & 6.816                     & 35.099 \\
FUnIE-GAN~\cite{islam2020fast}    & 0.386                      & 26.220                     & 0.792                     & 2.971                     & 0.212                     & 6.892                     & \textcolor{green}{30.912} \\
FUnIE-GAN-UP~\cite{islam2020fast}     & 0.600                      & 25.224                     & 0.788                     & 2.935                     & 0.246                     & 6.853                     & 34.070 \\
Deep SESR~\cite{islam2020simultaneous}     & 0.325                      & 27.081                     & 0.803                     & \textcolor{blue}{3.099}    & 0.206                     & \textcolor{red}{7.051}     & 35.179 \\
DWNet~\cite{sharma2023wavelength}         & \textcolor{green}{0.276}    & \textcolor{green}{28.654}   & \textcolor{green}{0.835}   & 3.042                     & \textcolor{blue}{0.173}    & \textcolor{blue}{7.051}    & \textcolor{blue}{30.856} \\
Ushape~\cite{peng2023u}      & 0.370                      & 26.822                     & 0.811                     & \textcolor{green}{3.052}    & 0.187                     & 6.843                     & 35.648 \\
Lit-Net  \cite{litnet}          & \textcolor{blue}{0.225}     & \textcolor{red}{29.477}    & \textcolor{blue}{0.851}    & 3.027                     & \textcolor{red}{0.169}     & 7.011                     & 32.109 \\
FUSION (Ours)             & \textcolor{red}{0.208}      & \textcolor{blue}{28.671}   & \textcolor{red}{0.862}    & \textcolor{red}{3.220}     & \textcolor{green}{0.174}   & \textcolor{green}{7.048}   & \textcolor{red}{29.547} \\
\hline
\end{tabular}
}
\label{tab:EUVP}
\end{table}

We first evaluate the performance of our proposed FUSION framework on three widely used underwater image datasets: UIEB~\cite{li2019underwater}, EUVP~\cite{islam2020fast}, and SUIM-E~\cite{qi2022sguie}. All images across these datasets are resized to a uniform resolution of 256×256 prior to training and evaluation. For training, we utilize the EUVP dataset, which contains 11,435 paired underwater images, while its test set consists of 515 image pairs of the same resolution. The UIEB dataset comprises 890 paired images, from which 800 are randomly selected for training, and the remaining 90 images are used for testing (following~\cite{li2019underwater}). The SUIM-E dataset includes 1,635 images, with 1,525 used for training and 110 for evaluation (following~\cite{litnet}).

To comprehensively assess the visual quality and perceptual fidelity of enhanced images, we compare our method against a range of state-of-the-art (SOTA) underwater image enhancement (UIE) approaches using both full-reference and no-reference metrics. These include Peak Signal-to-Noise Ratio (PSNR), Structural Similarity Index (SSIM), and Learned Perceptual Image Patch Similarity (LPIPS), along with perceptual quality measures such as the Underwater Image Quality Measure (UIQM), Underwater Image Sharpness Measure (UISM), and Blind/Referenceless Image Spatial Quality Evaluator (BRISQUE). Tables~\ref{tab:UIEB} and \ref{tab:EUVP} present a detailed comparison of the quantitative results on the UIEB and EUVP datasets, respectively.

\subsection{Comparison with State-of-the-Art}

\begin{figure}
    \centering
    \includegraphics[width=\columnwidth]{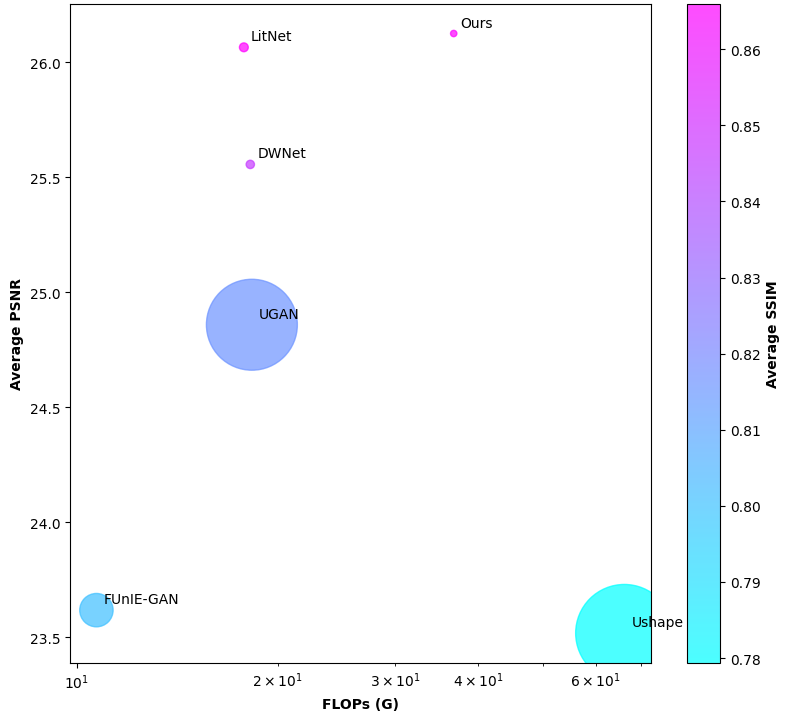}
    \caption{Bubble chart comparing the trade-off between average PSNR and GFLOPs for various UIE models}
    \label{fig:psnr_flops}
\end{figure}

We present a quantitative and quantitative evaluation demonstrating that FUSION consistently outperforms competing methods across all evaluated metrics, achieving state-of-the-art results. In particular, on the UIEB test set (Table~\ref{tab:UIEB}), FUSION achieves a PSNR of 23.717 dB and an SSIM of 0.883, indicating a very high reconstruction fidelity and structural similarity. It also records the lowest LPIPS score (0.112), reflecting superior perceptual quality and detail preservation. We observe similar trends on the EUVP dataset (Table~\ref{tab:EUVP}), where FUSION attains a PSNR of 28.671 dB and the highest SSIM value of 0.862, alongside a low LPIPS score (0.174). Figure~\ref{fig:psnr_flops} depicts a bubble chart illustrating the trade-off between average PSNR and GFLOPs for various models, further validating the balance between efficiency and effectiveness of our approach.
% These results collectively underscore the effectiveness of our dual-domain framework in enhancing underwater images by balancing local detail restoration with global color correction.

\begin{table}[ht]
\centering
\caption{Comparison with the model parameters and GFLOPs of SOTA models at an input size of $256 \times 256$. Lower is better. First best is in \textcolor{red}{red}, second best in \textcolor{blue}{blue}.}
\tiny
\resizebox{\columnwidth}{!}{
\begin{tabular}{lcc}
\hline
\textbf{Method} & \textbf{Parameters (M)} & \textbf{FLOPs (G)} \\
\hline
WaterNet~\cite{waternet}   & 24.8                   & 193.7                    \\
UGAN~\cite{fabbri2018enhancing}       & 57.17                  & 18.3                     \\
FUnIE-GAN~\cite{islam2020fast}  & 7.71                   & \textcolor{red}{10.7}    \\
Ucolor~\cite{Ucolor}     & 157.4                  & 443.9                    \\
SGUIE-Net~\cite{qi2022sguie}  & 18.55                  & 123.5                    \\
DWNet~\cite{sharma2023wavelength}      & \textcolor{blue}{0.48} & 18.2                     \\
Ushape~\cite{peng2023u}     & 65.6                   & 66.2                     \\
LitNet~\cite{litnet}     & 0.54                   & \textcolor{blue}{17.8}   \\
\rowcolor[gray]{0.8}
Ours       & \textcolor{red}{0.28}  & 36.73                    \\
\hline
\end{tabular}
}
\label{tab:params_flops}
\end{table}

We evaluate the visual quality of our FUSION framework through qualitative comparisons. Figures~\ref{fig:visual_UIEB} and \ref{fig:visual_EUVP} show enhancement results for the UIEB and EUVP datasets alongside outputs from state-of-the-art methods. FUSION recovers finer structural details and preserves subtle textures, restoring balanced color distributions and improving contrast to mitigate underwater distortions like color casts and low visibility. UIEB and EUVP results (Figure~\ref{fig:visual_UIEB}) enhance natural hues and recover important scene details better than competing methods.

\begin{table*}[!ht]
\centering
\caption{Ablation performance on UIEB.}
\resizebox{\textwidth}{!}{%
\begin{tabular}{lcccccccccc}
\hline
\textbf{Configuration} & \textbf{Freq. Attn} & \textbf{Freq. Branch} & \textbf{Freq. Fusion} & \textbf{Chan. Calib} & \textbf{Local Attn} & \textbf{Global Attn} & \textbf{UIQM} & \textbf{UISM} & \textbf{LPIPS} & \textbf{BRISQUE} \\
\hline
\rowcolor[gray]{0.8}
Full Model (FUSION) & \checkmark & \checkmark & \checkmark & \checkmark & \checkmark & \checkmark & 3.414 & 7.429 & 0.112 & 23.19 \\
No frequency attention & \xmark & \checkmark & \checkmark & \checkmark & \checkmark & \checkmark & 2.978 & 7.235 & 0.153 & 24.81 \\
No Frequency Branch & \checkmark & \xmark & \checkmark & \checkmark & \checkmark & \checkmark & 2.903 & 6.606 & 0.231 & 27.25 \\
No Frequency Guided Fusion & \checkmark & \checkmark & \xmark & \checkmark & \checkmark & \checkmark & 2.961 & 6.821 & 0.202 & 26.34 \\
No Channel Calibration & \checkmark & \checkmark & \checkmark & \xmark & \checkmark & \checkmark & 2.827 & 6.751 & 0.214 & 26.68 \\
No Local Attention & \checkmark & \checkmark & \checkmark & \checkmark & \xmark & \checkmark & 3.005 & 7.102 & 0.169 & 25.22 \\
No Global Attention & \checkmark & \checkmark & \checkmark & \checkmark & \checkmark & \xmark & 3.000 & 7.268 & 0.148 & 24.37 \\
Spatial Only & \xmark & \xmark & \xmark & \checkmark & \checkmark & \checkmark & 2.896 & 6.660 & 0.225 & 26.91 \\
Minimal Model & \xmark & \xmark & \xmark & \xmark & \xmark & \xmark & 2.720 & 6.410 & 0.258 & 28.43 \\
\hline
\end{tabular}%
}
\label{tab:ablation_uieb}
\end{table*}

\begin{table*}[ht]
\centering
\caption{Ablation Study Results on the EUVP Dataset}
\label{tab:ablation_euvp}
\resizebox{\textwidth}{!}{%
\begin{tabular}{lcccccccccc}
\toprule
\textbf{Configuration} & \textbf{Freq. Attn} & \textbf{Freq. Branch} & \textbf{Freq. Fusion} & \textbf{Chan. Calib} & \textbf{Local Attn} & \textbf{Global Attn} & \textbf{UIQM} & \textbf{UISM} & \textbf{LPIPS} & \textbf{BRISQUE} \\
\midrule
\rowcolor[gray]{0.8}
Full Model (FUSION)            & \checkmark & \checkmark & \checkmark & \checkmark & \checkmark & \checkmark & 3.220 & 7.048 & 0.174 & 29.547 \\
No Frequency Attention        & \xmark     & \checkmark & \checkmark & \checkmark & \checkmark & \checkmark & 2.839 & 6.118 & 0.227 & 34.21 \\
No Frequency Branch           & \checkmark & \xmark     & \checkmark & \checkmark & \checkmark & \checkmark & 2.674 & 5.709 & 0.249 & 35.68 \\
No Frequency Guided Fusion    & \checkmark & \checkmark & \xmark     & \checkmark & \checkmark & \checkmark & 2.665 & 5.744 & 0.247 & 35.53 \\
No Channel Calibration        & \checkmark & \checkmark & \checkmark & \xmark     & \checkmark & \checkmark & 2.640 & 5.646 & 0.252 & 35.89 \\
No Local Attention            & \checkmark & \checkmark & \checkmark & \checkmark & \xmark     & \checkmark & 2.538 & 6.222 & 0.232 & 34.51 \\
No Global Attention           & \checkmark & \checkmark & \checkmark & \checkmark & \checkmark & \xmark     & 2.640 & 6.392 & 0.224 & 33.92 \\
Spatial Only                  & \xmark     & \xmark     & \xmark     & \checkmark & \checkmark & \checkmark & 2.373 & 5.557 & 0.261 & 36.43 \\
Minimal Model                 & \xmark     & \xmark     & \xmark     & \xmark     & \xmark     & \xmark     & 2.106 & 5.553 & 0.278 & 37.21 \\
\bottomrule
\end{tabular}
}
\end{table*}

In addition to quantitative performance, we also assess the efficiency of our approach. Table~\ref{tab:params_flops} summarizes the model parameters and GFLOPs for our method compared to other leading UIE models at an input size of 256$\times$256. Notably, FUSION achieves superior enhancement results with a significantly lower number of parameters (0.28M) and competitive GFLOPs (36.73), justifying its potential for deployment in real-time and resource-constrained settings.

\subsection{Ablation Study}

\textbf{Quantitative Analysis.} From the ablation studies across UIEB and EUVP, it is evident that each architectural component contributes meaningfully to overall performance. Removing frequency attention, branch, or guided fusion consistently leads to notable degradation in perceptual quality (higher LPIPS, lower UIQM and UISM), affirming the critical role of frequency-aware design in FUSION. Similarly, channel calibration and attention blocks - both local and global - also drive significant gains, especially in structural sharpness and perceptual realism. Interestingly, global attention appears to be particularly vital in retaining fine-grained global coherence, while local attention improves texture fidelity. Models stripped of frequency modules or reduced to spatial-only designs suffer from reduced enhancement quality, confirming the synergy between spectral and spatial representations in underwater image enhancement.

\textbf{Hardware Efficiency.} Beyond accuracy, FUSION maintains competitive inference efficiency, showcasing a balanced trade-off between performance and resource footprint. The full model runs at 128.68 ms with just 36.73 GFLOPs, which is notably efficient given its multi-branch design. Ablating the frequency branch or removing attention mechanisms reduces inference time - e.g., down to 75.87 ms without local attention - but at the cost of performance. While the minimal model is fastest at 18.49 ms, it offers the weakest performance, backing the need for our architectural complexity to achieve enhancement fidelity. Overall, FUSION demonstrates that strategic architectural additions, particularly those exploiting frequency and attention cues, yield meaningful gains without sacrificing deployability in real-time or resource-limited scenarios.

\section{Conclusion}
We propose FUSION (Frequency-guided Underwater Spatial Image recOnstructioN), a novel dual-domain framework that combines multi-scale spatial feature extraction with FFT-based frequency processing for underwater image enhancement. Leveraging adaptive attention, FUSION effectively addresses complex degradations in underwater scenes. Extensive evaluations on UIEB, EUVP, and SUIM-E show superior performance across PSNR, SSIM, LPIPS, UIQM, UISM, and BRISQUE metrics. FUSION also offers a strong balance between quality and efficiency, making it ideal for real-time use on AUVs.

% Overall, FUSION provides an accurate, efficient, and a SOTA solution for underwater image enhancement, advancing over all the current state-of-the-art methods.

{
    \small
    \bibliographystyle{ieeenat_fullname}
    \bibliography{main}
}
\clearpage
\setcounter{page}{1}
\maketitlesupplementary

\subsection*{Additional Extended Methodology}
In this section, we expand upon the mathematical foundations of our framework, detailing the operations performed in both the spatial and frequency domains, as well as their fusion and calibration.

\textbf{1. Spatial Domain Processing:}  
For an input image \( D^{h \times w \times 3} \), each color channel \( D_i \) (with \( i \in \{R, G, B\} \)) is processed independently. The initial multi-scale feature extraction is given by:
\begin{equation}
    f^1_i = \Phi_i(D_i),
\end{equation}
where \( \Phi_i(\cdot) \) denotes convolutional operations with kernel sizes \(3 \times 3\) (for \(R\)), \(5 \times 5\) (for \(G\)), and \(7 \times 7\) (for \(B\)). To enhance these features, a two-stage attention mechanism is applied.

First, channel attention is computed as:
\begin{equation}
    W_{\text{channel}} = \sigma\Big( \mathbf{W}_2 \cdot \phi\big(\mathbf{W}_1 \cdot g(f^1_i)\big) \Big),
\end{equation}
where \( g(f^1_i) \) denotes global average pooling, \( \phi(\cdot) \) is a ReLU activation, and \( \mathbf{W}_1 \) and \( \mathbf{W}_2 \) are learnable weight matrices. The feature map is then scaled element-wise:
\begin{equation}
    f_{\text{channel-att}} = W_{\text{channel}} \odot f^1_i.
\end{equation}
Next, spatial attention is defined by:
\begin{equation}
    W_{\text{spatial}} = \sigma\Big( \psi\big( [ \mathcal{P}_{avg}(f_{\text{channel-att}}); \mathcal{P}_{max}(f_{\text{channel-att}}) ] \big) \Big),
\end{equation}
where \( \mathcal{P}_{avg} \) and \( \mathcal{P}_{max} \) denote average and max pooling, respectively, and \( \psi(\cdot) \) is a convolutional mapping. The refined spatial features are obtained as:
\begin{equation}
    f^2_i = W_{\text{spatial}} \odot \big( W_{\text{channel}} \odot f^1_i \big).
\end{equation}
Finally, a residual connection ensures low-level features are preserved:
\begin{equation}
    f^3_i = f^2_i + f^1_i, \quad \forall i \in \{R, G, B\}.
\end{equation}

\textbf{2. Frequency Domain Processing:}  
Each channel \( D_i \) is transformed into the frequency domain using the 2D Fast Fourier Transform (FFT):
\begin{equation}
    F_i(u,v) = \sum_{x=0}^{h-1}\sum_{y=0}^{w-1} D_i(x,y) \, e^{-j2\pi\left(\frac{ux}{h}+\frac{vy}{w}\right)}.
\end{equation}
The magnitude of the frequency representation is computed as:
\begin{equation}
    |F_i(u,v)| = \sqrt{\text{Re}(F_i(u,v))^2 + \text{Im}(F_i(u,v))^2}.
\end{equation}
To refine the magnitude features, we perform a linear transformation:
\begin{equation}
    \hat{F}_i = W_2 \cdot \phi\Big( W_1 \cdot |F_i| \Big),
\end{equation}
followed by frequency attention:
\begin{equation}
    W_{\text{freq}} = \sigma\Big( W_4 \cdot \phi\big( W_3 \cdot \bar{F}_i \big) \Big), \quad \bar{F}_i = \frac{1}{hw}\sum_{u,v} |F_i(u,v)|.
\end{equation}
The refined magnitude is:
\begin{equation}
    |F_i|_{\text{refined}} = W_{\text{freq}} \odot |F_i|.
\end{equation}
The phase information \( \Theta_i(u,v) \) is preserved as:
\begin{equation}
    \Theta_i(u,v) = \arctan\Big(\frac{\text{Im}(F_i(u,v))}{\text{Re}(F_i(u,v))}\Big),
\end{equation}
and the refined complex representation is reconstructed by:
\begin{equation}
    F'_i(u,v) = |F_i|_{\text{refined}} \cdot e^{j\Theta_i(u,v)}.
\end{equation}
Finally, the inverse FFT recovers the spatial features:
\begin{equation}
    f_{\text{freq},i}(x,y) = \frac{1}{hw} \sum_{u=0}^{h-1}\sum_{v=0}^{w-1} F'_i(u,v) \, e^{j2\pi\left(\frac{ux}{h}+\frac{vy}{w}\right)}.
\end{equation}

\textbf{3. Frequency Guided Fusion (FGF):}  
The spatial features \( f^3_i \) and frequency features \( f_{\text{freq},i} \) are fused to form a unified representation:
\begin{equation}
    f_{\text{concat},i} = \text{Concat}\Big( f^3_i,\, f_{\text{freq},i} \Big).
\end{equation}
A convolutional layer then integrates these features:
\begin{equation}
    f_{\text{fused},i} = \phi\Big( W_i * f_{\text{concat},i} + b_i \Big),
\end{equation}
where \( * \) denotes convolution and \( b_i \) is the bias term.

\textbf{4. Inter-Channel Fusion and Channel Calibration:}  
Fused representations from the three channels are concatenated:
\begin{equation}
    f_{\text{all}} = \text{Concat}\Big( f_{\text{fused},R},\, f_{\text{fused},G},\, f_{\text{fused},B} \Big).
\end{equation}
This aggregated feature is projected into a higher-dimensional space:
\begin{equation}
    f_d = \phi\Big( \mathcal{T}_d(f_{\text{all}}) \Big),
\end{equation}
and further integrated with frequency features through a learned transformation:
\begin{equation}
    f_{\text{fusion}} = \phi\Big( \mathcal{T}_f(f_d, f_{\text{freq}}) \Big).
\end{equation}
A global attention mechanism refines this fused representation:
\begin{equation}
    f_{\text{attn}} = \mathcal{A}\Big( f_{\text{fusion}},\, f_{\text{all}} \Big),
\end{equation}
followed by the reconstruction of a preliminary enhanced image:
\begin{equation}
    E = \phi\Big( \mathcal{T}_e(f_{\text{attn}}) \Big).
\end{equation}
Finally, adaptive channel calibration is performed:
\begin{equation}
    W_{\text{calibration}} = \sigma\Big( W_2 \cdot \phi\big( W_1 \cdot g(E) \big) \Big),
\end{equation}
\begin{equation}
    E_{\text{final}} = E \odot W_{\text{calibration}},
\end{equation}
ensuring that the final enhanced image \( E_{\text{final}} \) exhibits balanced color distributions and preserved structural details.

\subsection*{Hardware and Training Details}
We run all our experiments on a NVIDIA Tesla P100 GPU (Pascal architecture) with 16 GB of HBM2 memory and 3,584 CUDA cores, delivering up to 9.3 TFLOPS of single-precision performance. Since our method focuses on lightweight design and real-time feasibility, testing on a GPU with minimal compute ensures efficiency without relying on heavy hardware. We also use automatic mixed-precision (AMP) to speed up training and reduce memory usage, making the process even more efficient.

\textbf{Training Settings:}  
We train our model using the Adam optimizer with a starting learning rate of \(2 \times 10^{-4}\), \(\beta_1 = 0.5\), and \(\beta_2 = 0.999\). Training runs for up to 1,000 epochs with a batch size of 4, but we use early stopping based on LPIPS. We choose LPIPS since it closely aligns with human perception, ensuring that the model focuses on producing visually improved underwater images.

\textbf{Computation-Related Info:}
\begin{verbatim}
===== Model Performance Report =====
GPU Memory Used: 50.46 MB
Peak GPU Memory: 260.34 MB
Inference Time: 1.8147 seconds
Estimated FPS: 0.55 frames per second
Total FLOPs: 18.41 GFLOPs
\end{verbatim}

\section*{Supplementary Results}+

\begin{figure}[!h]
    \centering
    \includegraphics[width=\columnwidth]{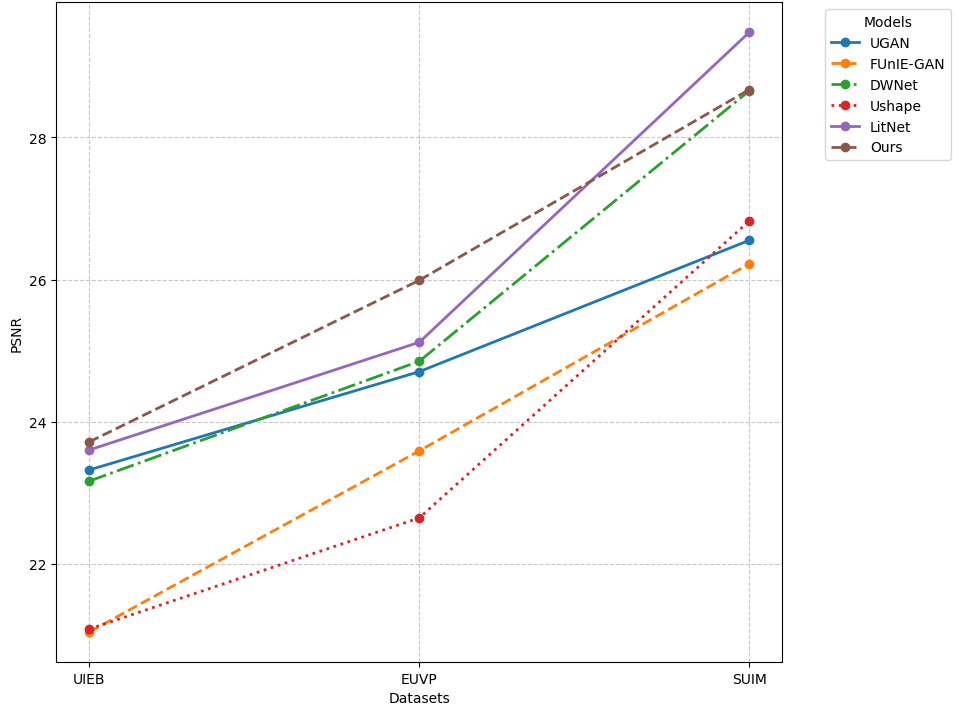}
    \caption{Line chart comparing PSNR values across the UIEB, EUVP, and SUIM-E datasets.}
    \label{fig:line_psnr}
\end{figure}

\begin{figure}[!h]
    \centering
    \includegraphics[width=\columnwidth]{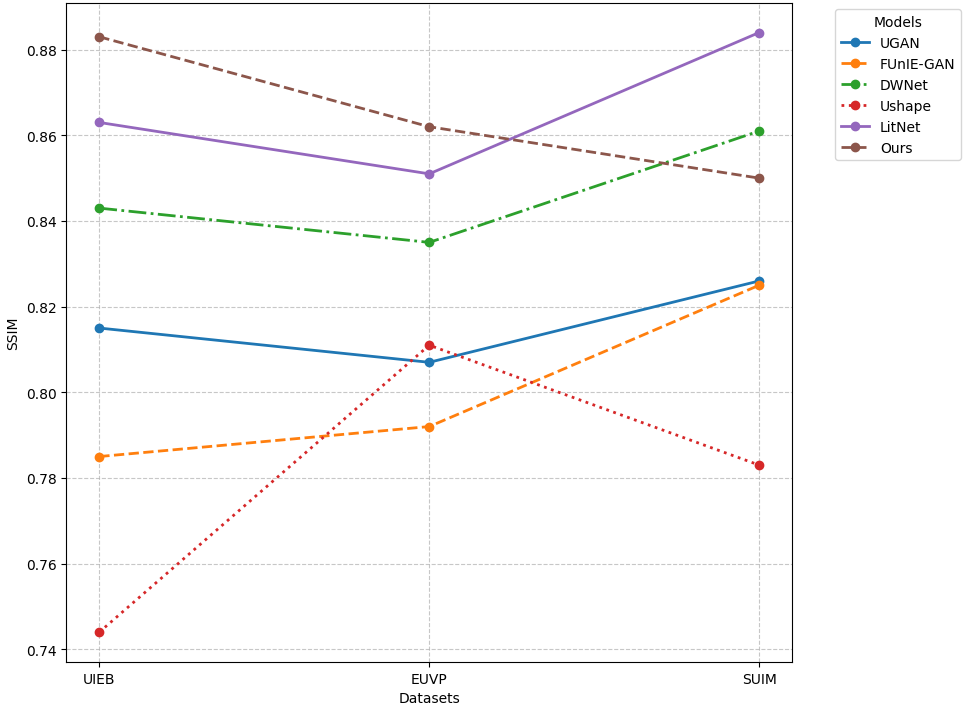}
    \caption{Line chart comparing SSIM values across the UIEB, EUVP, and SUIM-E datasets.}
    \label{fig:line_ssim}
\end{figure}

In these supplementary results, we provide additional quantitative and qualitative visualizations to further illustrate the performance and efficiency of our proposed FUSION framework. In addition to the primary metrics presented in the main paper, these supplementary results include detailed ablation studies, bar plots comparing quality metrics across the UIEB, EUVP, and SUIM-E datasets, as well as extended efficiency analyses. These visualizations are closely tied to the mathematical formulations described in Section~\ref{sec:proposedMethod} and underscore the importance of our dual-domain processing.

\begin{table}[h]
\centering
\caption{Evaluation on SUIM-E test set with the best-published works for UIE. First, second, and third best performances are represented in \textcolor{red}{red}, \textcolor{blue}{blue}, and \textcolor{green}{green} colors, respectively. ↓ indicates lower is better.}
\resizebox{\columnwidth}{!}{
\begin{tabular}{lcccccc}
\hline
Method           & PSNR              & SSIM                       & LPIPS↓                           & UIQM                          & UISM                          & BRISQUE↓                      \\
\hline
UDCP~\cite{drews2013transmission}    & 12.074            & 0.513                      & 0.270                            & 1.648                        & \textcolor{blue}{7.537}       & 22.788                        \\
GBdehaze~\cite{li2016single}         & 14.339            & 0.599                      & 0.355                            & 2.255                        & 7.400                        & 20.175                        \\
IBLA~\cite{peng2017underwater}       & 18.024            & 0.685                      & 0.209                            & 1.826                        & 7.341                        & 20.957                        \\
ULAP~\cite{song2018rapid}            & 19.148            & 0.744                      & 0.231                            & 2.115                        & \textcolor{green}{7.475}      & 21.250                        \\
CBF~\cite{ancuti2017color}           & 20.395            & 0.834                      & 0.194                            & \textcolor{blue}{3.003}       & 7.360                        & 21.115                        \\
UGAN~\cite{fabbri2018enhancing}      & 24.704            & 0.826                      & 0.190                            & 2.894                        & 7.175                        & 20.288                        \\
UGAN-P~\cite{fabbri2018enhancing}    & 25.050            & 0.827                      & 0.188                            & 2.901                        & 7.184                        & \textcolor{blue}{18.768}      \\
FUnIE-GAN~\cite{islam2020fast}       & 23.590            & 0.825                      & 0.189                            & \textcolor{green}{2.918}      & 7.121                        & 22.560                        \\
SGUIE-Net~\cite{qi2022sguie}         & \textcolor{blue}{25.987}  & \textcolor{green}{0.857}      & 0.153                            & 2.637                        & 7.090                        & 25.927                        \\
DWNet~\cite{sharma2023wavelength}    & 24.850            & \textcolor{blue}{0.861}      & \textcolor{green}{0.133}           & 2.707                        & 7.381                        & 20.757                        \\
Ushape~\cite{peng2023u}              & 22.647            & 0.783                      & 0.213                            & 2.873                        & 7.061                        & 22.876                        
                      \\
Lit-Net \cite{litnet}           & \textcolor{green}{25.117} & \textcolor{red}{0.884}       & \textcolor{red}{0.118}             & 2.918                        & 7.368                        & \textcolor{green}{19.602}     \\
FUSION (Ours)             & \textcolor{red}{25.989}   & \textcolor{blue}{0.850}      & \textcolor{blue}{0.118}            & \textcolor{red}{3.183}        & \textcolor{red}{7.679}        & \textcolor{red}{18.655}       \\
\hline
\end{tabular}
}
\label{tab:SUIME}
\end{table}

\begin{figure*}[!h]
    \centering
    \includegraphics[width=0.7\linewidth]{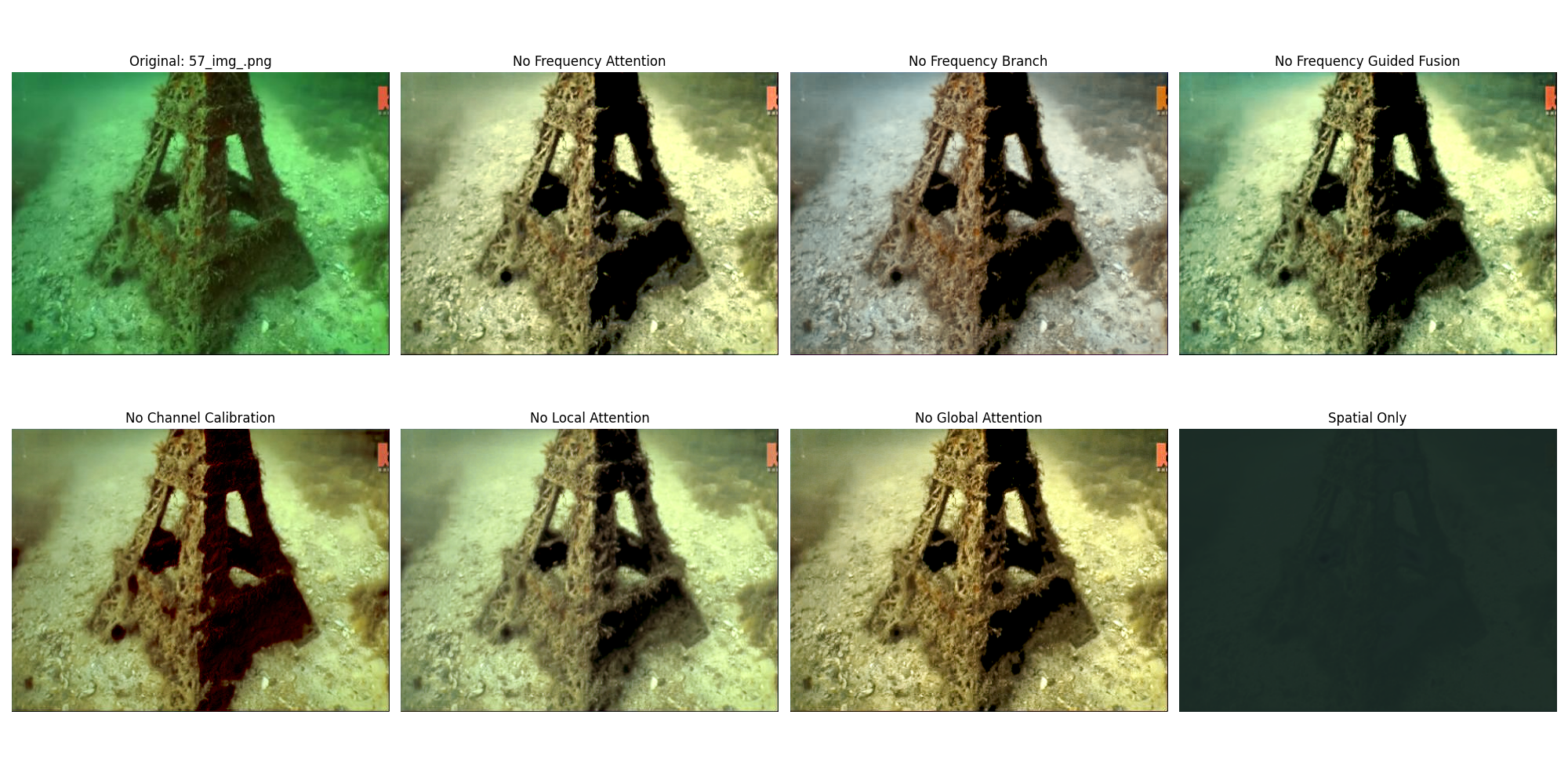}
    \caption{Ablation Study Visual Comparisons. This figure displays the enhancement results for a representative underwater image using our model with various component ablations: Original, No Frequency Attention, No Frequency Branch, No Frequency Guided Fusion, No Channel Calibration, No Local Attention, No Global Attention, and Spatial Only. The qualitative differences underscore the contribution of each module to the final enhancement quality.}
    \label{fig:ablation_comparison}
\end{figure*}

The above figure (\ref{fig:ablation_comparison}) illustrates the impact of ablating key components of our proposed model. We observe that removing individual modules results in visible degradations, which verifies the necessity of each part for achieving optimal performance.

\begin{figure*}[!h]
    \centering
    \includegraphics[width=\linewidth]{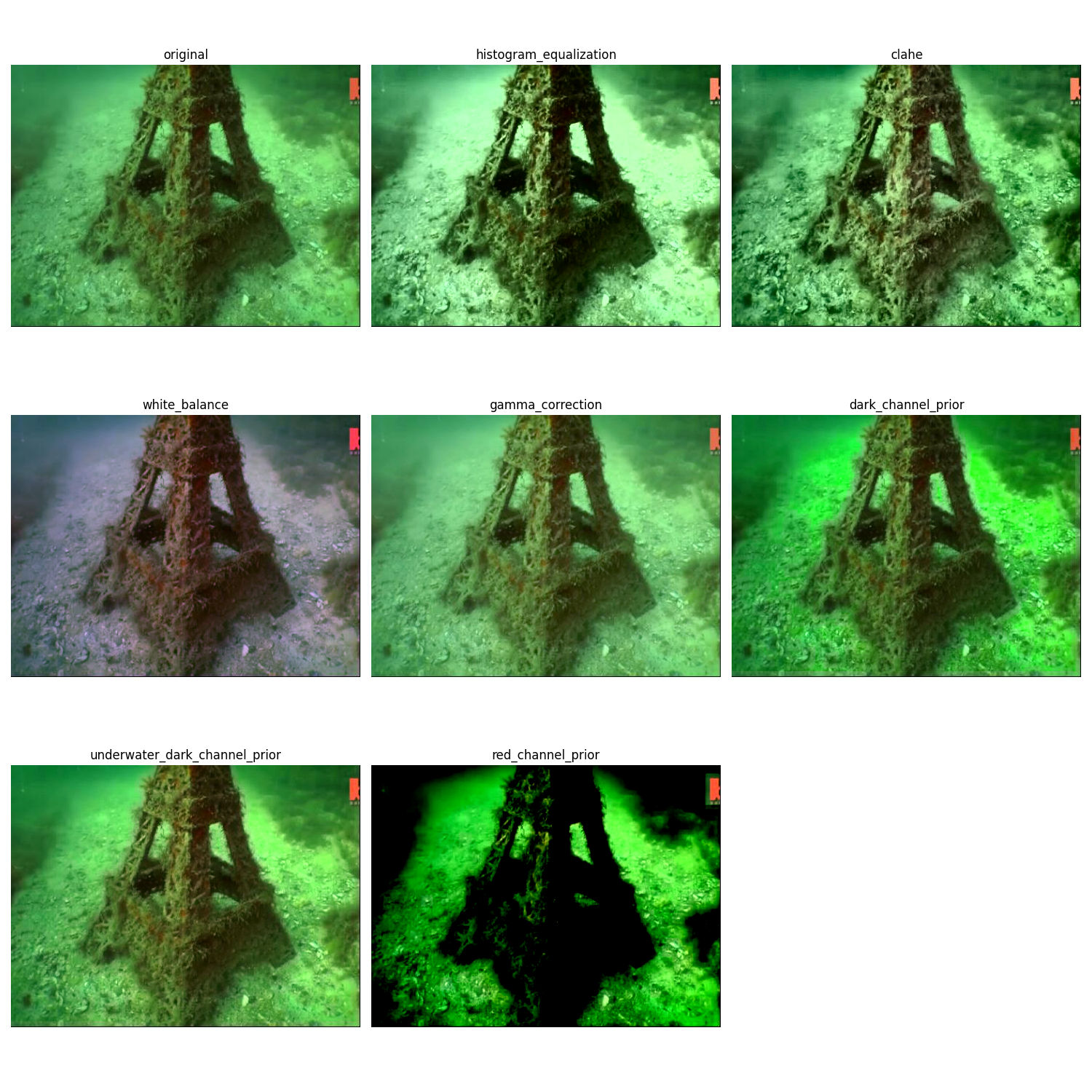}
    \caption{Comparison with Traditional Image Processing Techniques. The figure compares the original underwater image with images processed by conventional methods: histogram equalization, CLAHE, white balance, gamma correction, dark channel prior, underwater dark channel prior, and red channel prior. These comparisons highlight the limitations of traditional methods relative to our approach.}
    \label{fig:traditional_comparison}
\end{figure*}

Figure~\ref{fig:traditional_comparison} presents a qualitative comparison between our FUSION framework and several traditional image processing techniques. Notably, while methods such as histogram equalization and dark channel priors provide some level of enhancement, they fall short of recovering natural color balance and structural details, as achieved by our method.

\begin{figure*}[!ht]
    \centering
    \includegraphics[width=\linewidth]{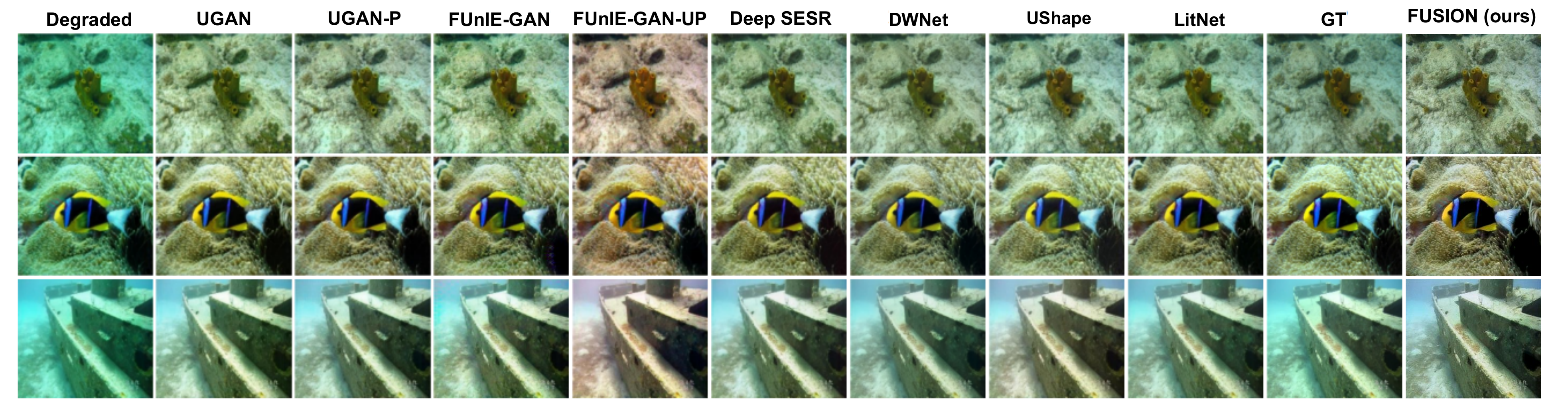}
    \caption{Visual comparisons on the SUIM-E dataset.}
    \label{fig:visual_SUIME}
\end{figure*}89

Figures~\ref{fig:line_psnr} and~\ref{fig:line_ssim} plot PSNR and SSIM values across the datasets. The PSNR chart shows FUSION consistently achieves higher reconstruction fidelity with elevated PSNR values. The SSIM chart reveals superior structural similarity compared to other approaches, even under challenging conditions. These plots highlight that FUSION enhances local details and color balance while preserving global image structure, reinforcing its effectiveness in underwater image enhancement tasks.

\begin{table*}[!ht]
\centering
\caption{Ablation performance on SUIM-E.}
\resizebox{\textwidth}{!}{%
\begin{tabular}{lcccccccccc}
\hline
Config & Freq. Attn & Freq. Branch & Freq. Fusion & Chan. Calib & Local Attn & Global Attn & UIQM & UISM & LPIPS & BRISQUE \\
\hline
Full Model (FUSION) & \checkmark & \checkmark & \checkmark & \checkmark & \checkmark & \checkmark & 3.183 & 7.679 & 0.118 & 18.655 \\
no\_frequency attention & \xmark & \checkmark & \checkmark & \checkmark & \checkmark & \checkmark & 2.626 & 5.832 & 0.242 & 23.91 \\
no\_frequency branch & \checkmark & \xmark & \checkmark & \checkmark & \checkmark & \checkmark & 2.806 & 5.674 & 0.285 & 25.73 \\
no\_frequency guided fusion & \checkmark & \checkmark & \xmark & \checkmark & \checkmark & \checkmark & 2.703 & 6.010 & 0.230 & 23.24 \\
no\_channel calibration & \checkmark & \checkmark & \checkmark & \xmark & \checkmark & \checkmark & 2.721 & 6.096 & 0.225 & 22.98 \\
no\_local attention & \checkmark & \checkmark & \checkmark & \checkmark & \xmark & \checkmark & 2.736 & 6.034 & 0.228 & 23.15 \\
no\_global attention & \checkmark & \checkmark & \checkmark & \checkmark & \checkmark & \xmark & 2.746 & 6.023 & 0.229 & 23.21 \\
spatial only & \xmark & \xmark & \xmark & \checkmark & \checkmark & \checkmark & 2.645 & 5.506 & 0.265 & 24.62 \\
minimal model & \xmark & \xmark & \xmark & \xmark & \xmark & \xmark & 2.445 & 5.149 & 0.292 & 26.18 \\
\hline
\end{tabular}%
}
\label{tab:ablation_suime}
\end{table*}

On the SUIM-E test set (Table~\ref{tab:SUIME}), our approach further confirms its robustness by achieving comparable PSNR, SSIM, and LPIPS scores. Additionally, FUSION exhibits favorable performance in perceptual quality metrics, with competitive UIQM, UISM, and BRISQUE scores across all datasets.

Our methodology leverages multi-scale convolutions, adaptive attention mechanisms, and frequency-domain transformations to address the complex degradations in underwater images. To offer deeper insight into our approach, we now provide additional mathematical details that further elaborate on the operations used in FUSION.

\begin{figure*}[!t]
    \centering
    \includegraphics[width=\linewidth]{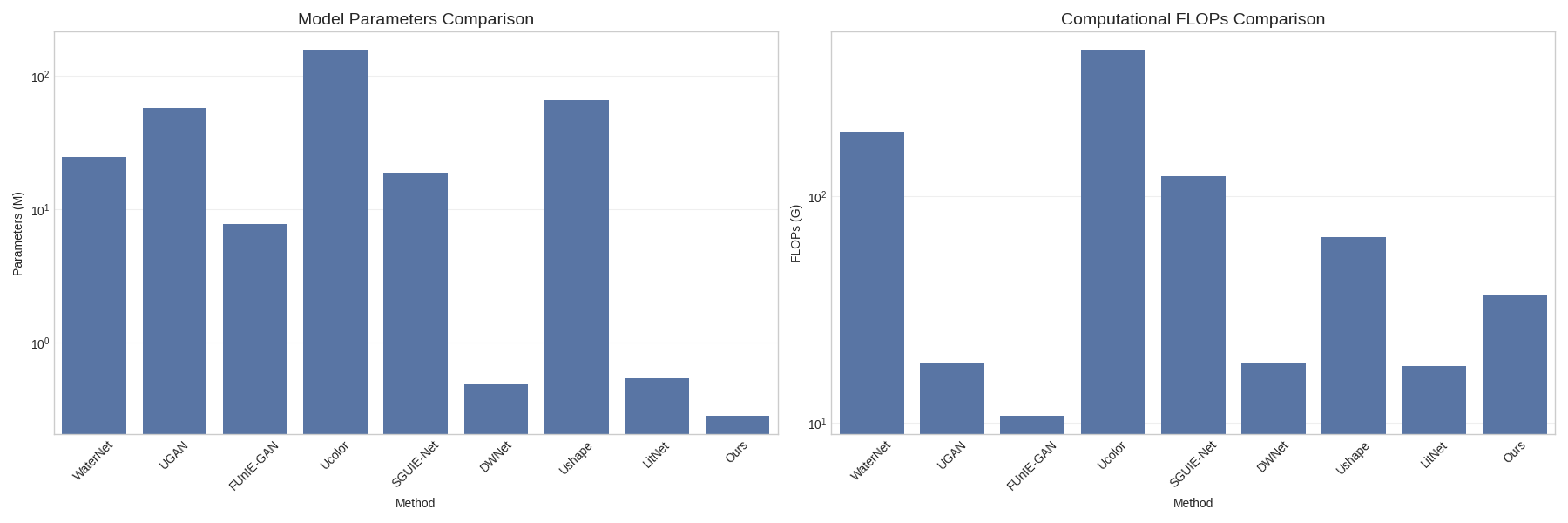}
    \caption{Side-by-side comparison of model parameters and GFLOPs for various UIE methods. FUSION achieves low computational cost without compromising enhancement performance.}
    \label{fig:side_efficiency}
\end{figure*}

\subsection*{Metric-wise Bar Plots}
\begin{figure*}[!t]
    \centering
    \subfloat[UIEB]{\includegraphics[width=0.32\textwidth]{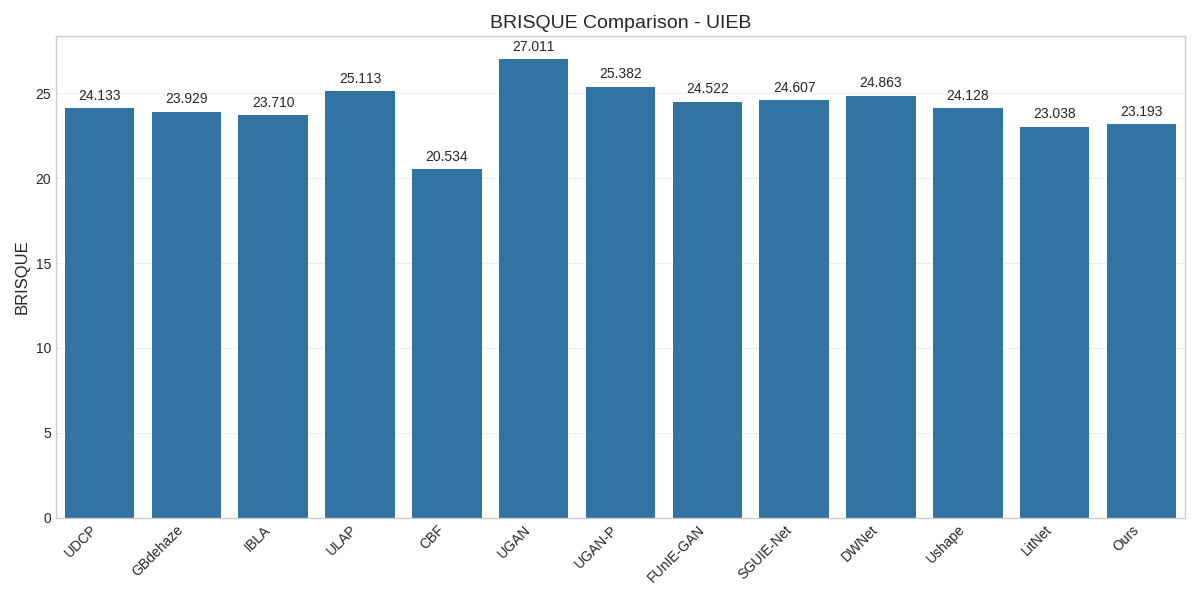}}
    \subfloat[EUVP]{\includegraphics[width=0.32\textwidth]{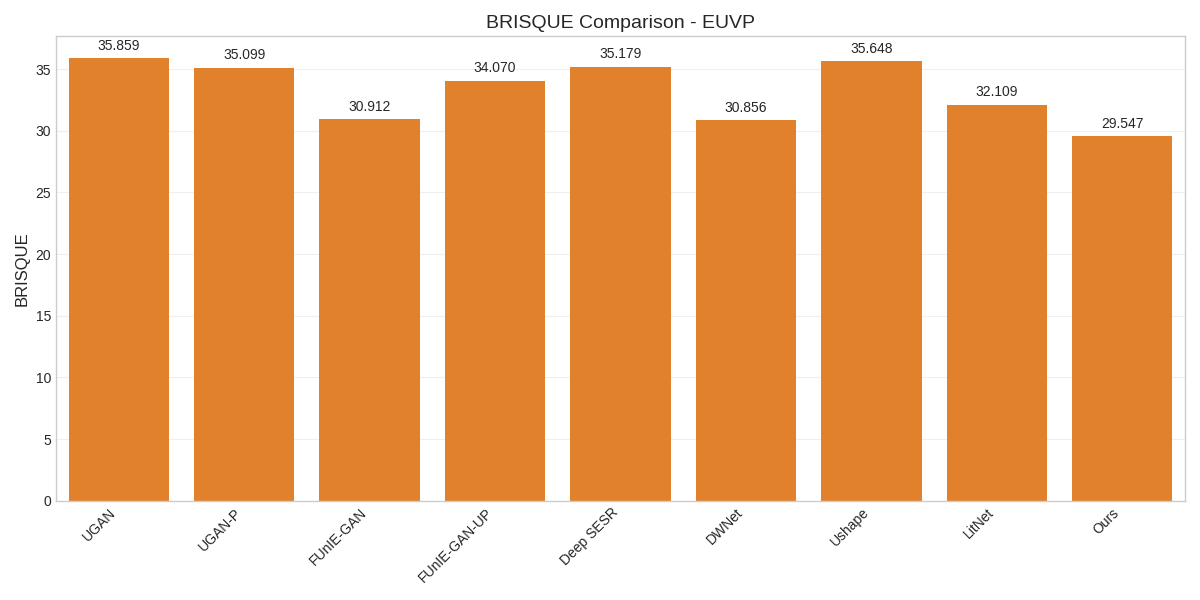}}
    \subfloat[SUIM-E]{\includegraphics[width=0.32\textwidth]{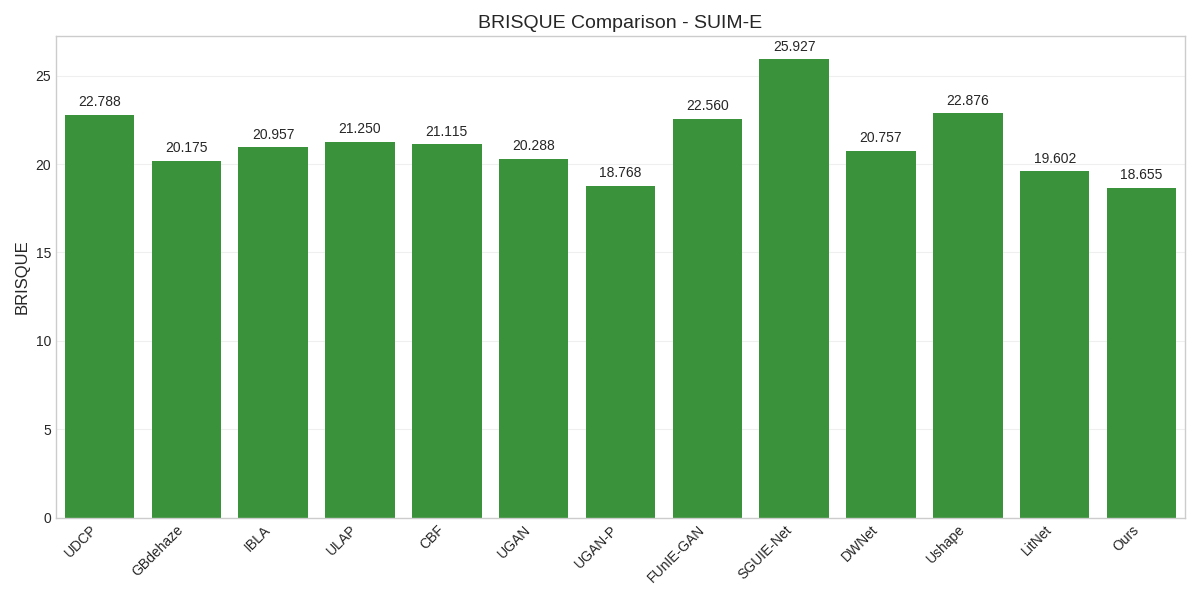}}
    \caption{Bar plots comparing BRISQUE scores (lower is better) across the UIEB, EUVP, and SUIM-E datasets. }
    \label{fig:bar_brisque}
\end{figure*}

\begin{figure*}[!t]
    \centering
    \subfloat[UIEB]{\includegraphics[width=0.32\textwidth]{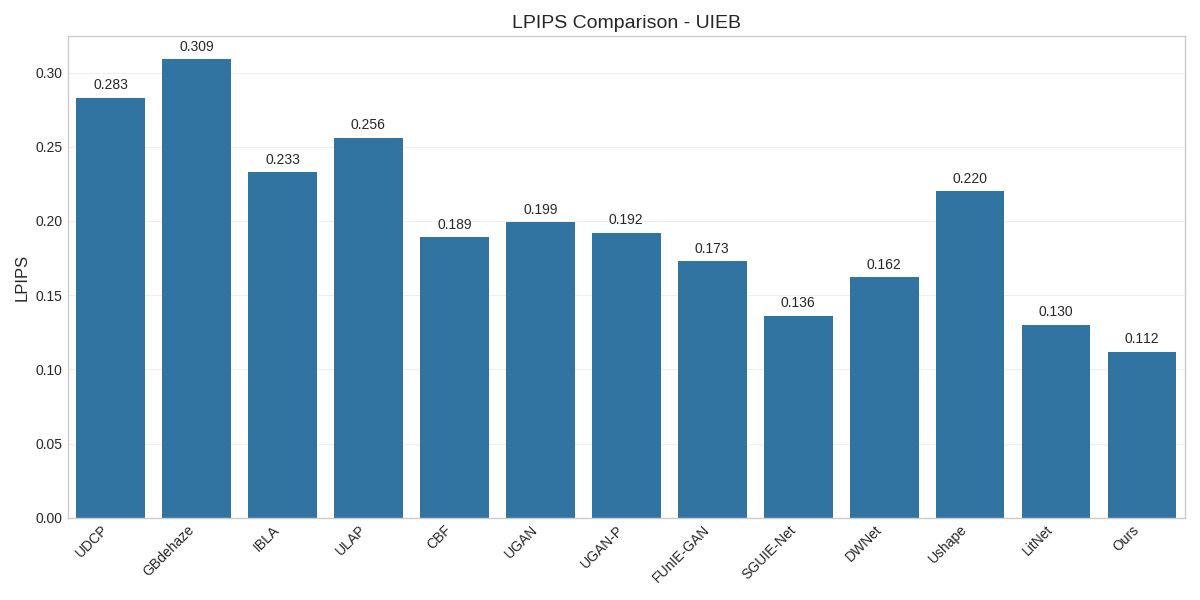}}
    \subfloat[EUVP]{\includegraphics[width=0.32\textwidth]{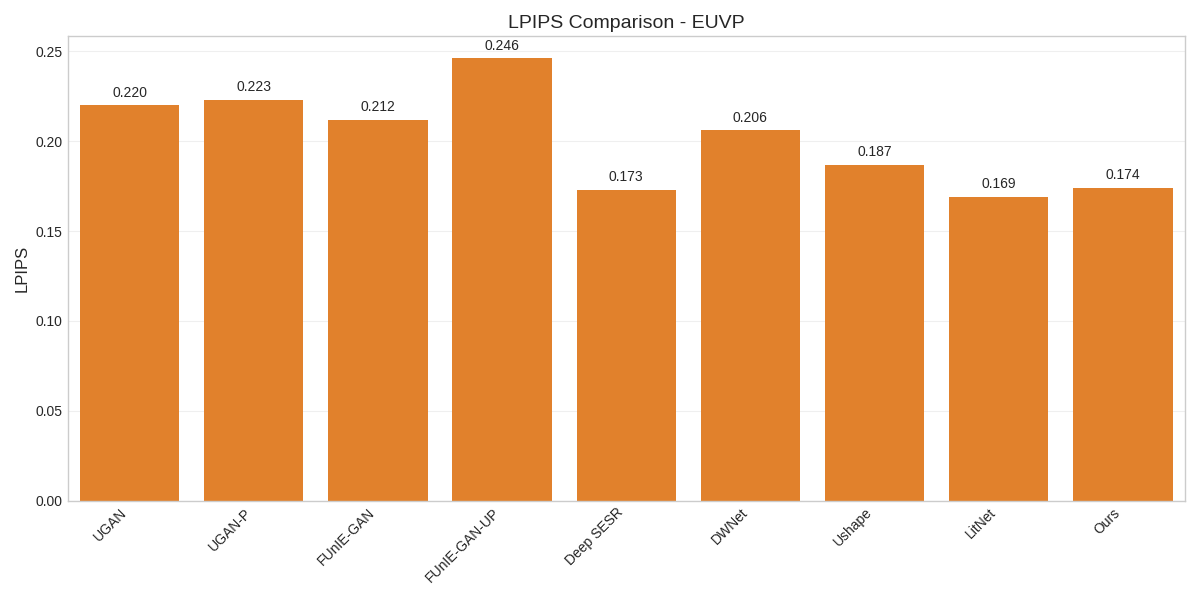}}
    \subfloat[SUIM-E]{\includegraphics[width=0.32\textwidth]{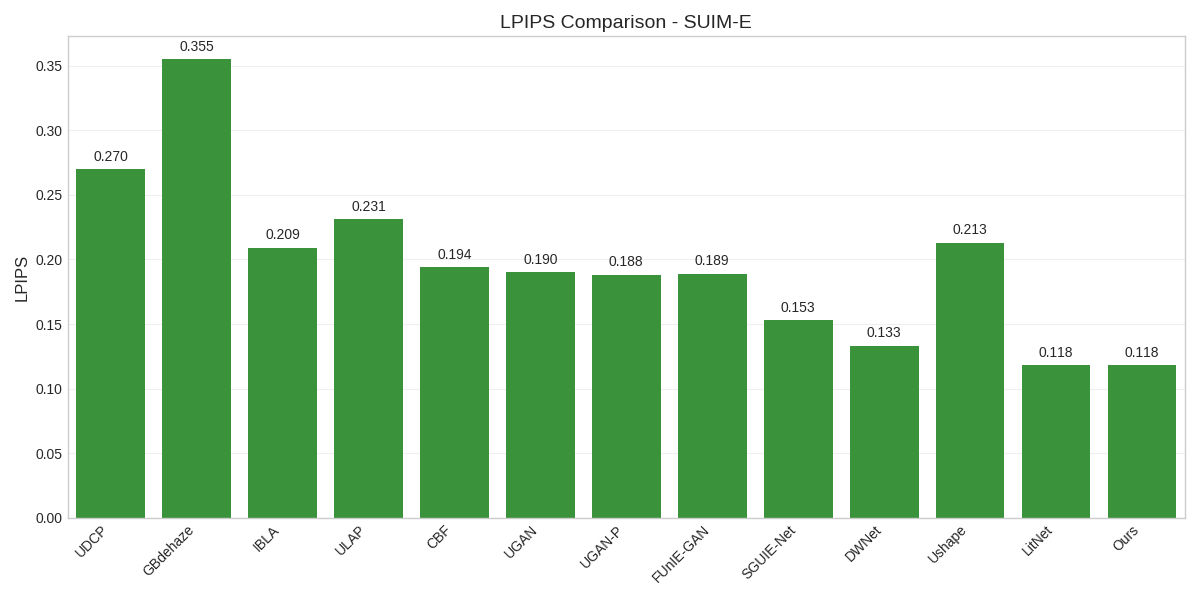}}
    \caption{Bar plots comparing LPIPS scores (lower is better) across the three datasets. Lower LPIPS values indicate that FUSION produces enhanced images that are perceptually closer to the ground truth.}
    \label{fig:bar_lpips}
\end{figure*}

\begin{figure*}[!t]
    \centering
    \subfloat[UIEB]{\includegraphics[width=0.32\textwidth]{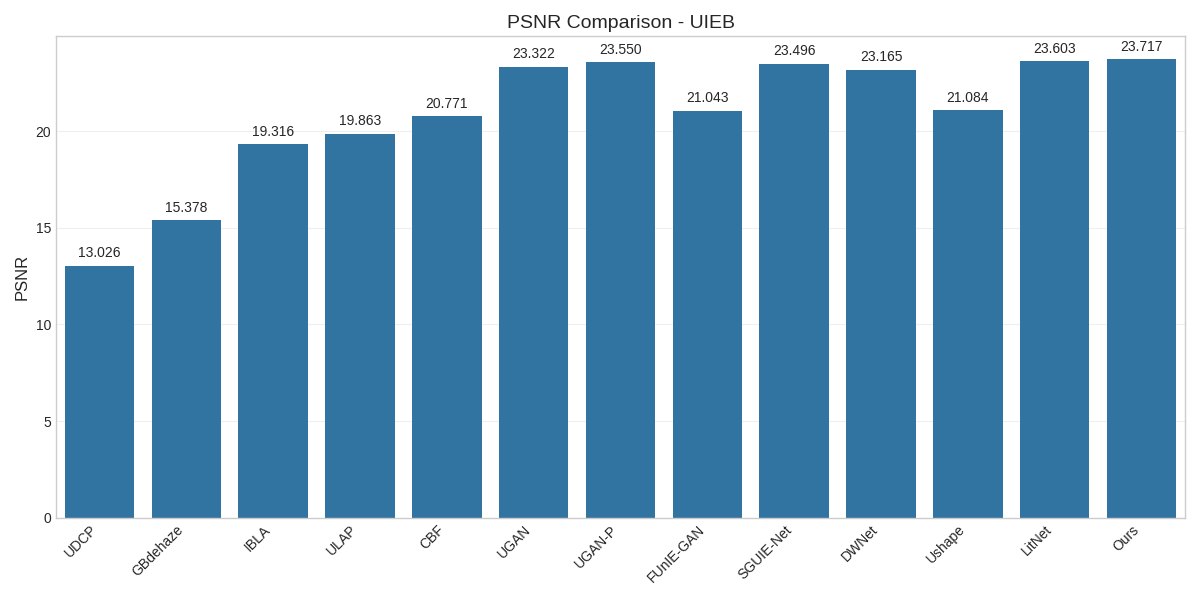}}
    \subfloat[EUVP]{\includegraphics[width=0.32\textwidth]{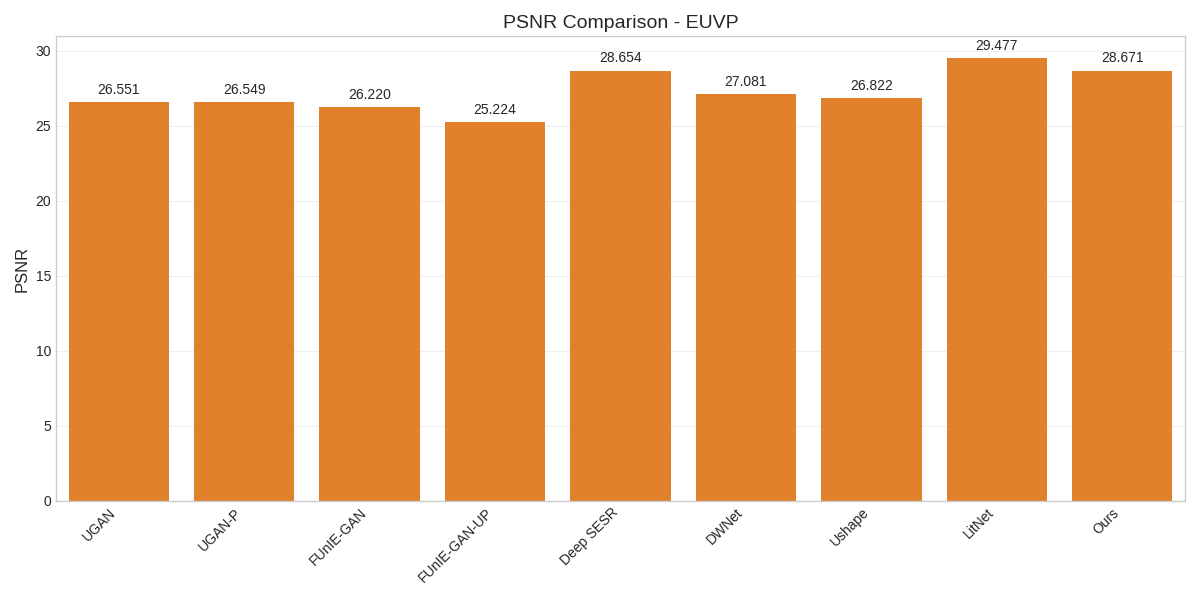}}
    \subfloat[SUIM-E]{\includegraphics[width=0.32\textwidth]{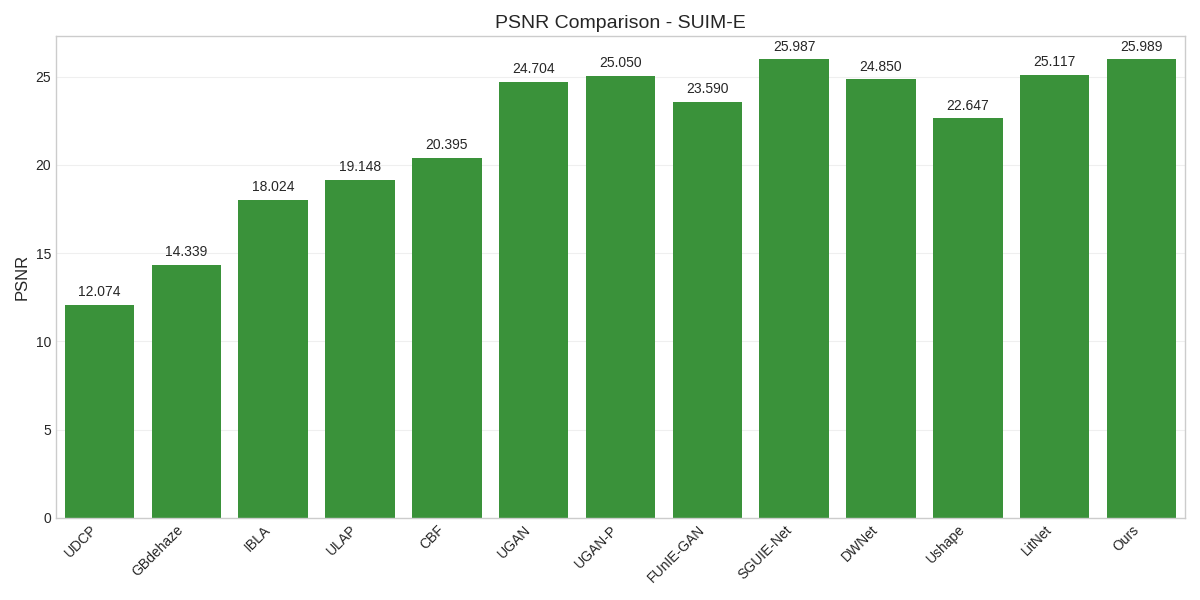}}
    \caption{Bar plots comparing PSNR values across the UIEB, EUVP, and SUIM-E datasets. Higher PSNR values achieved by FUSION indicate its reconstruction fidelity.}
    \label{fig:bar_psnr}
\end{figure*}

\begin{figure*}[!t]
    \centering
    \subfloat[UIEB]{\includegraphics[width=0.32\textwidth]{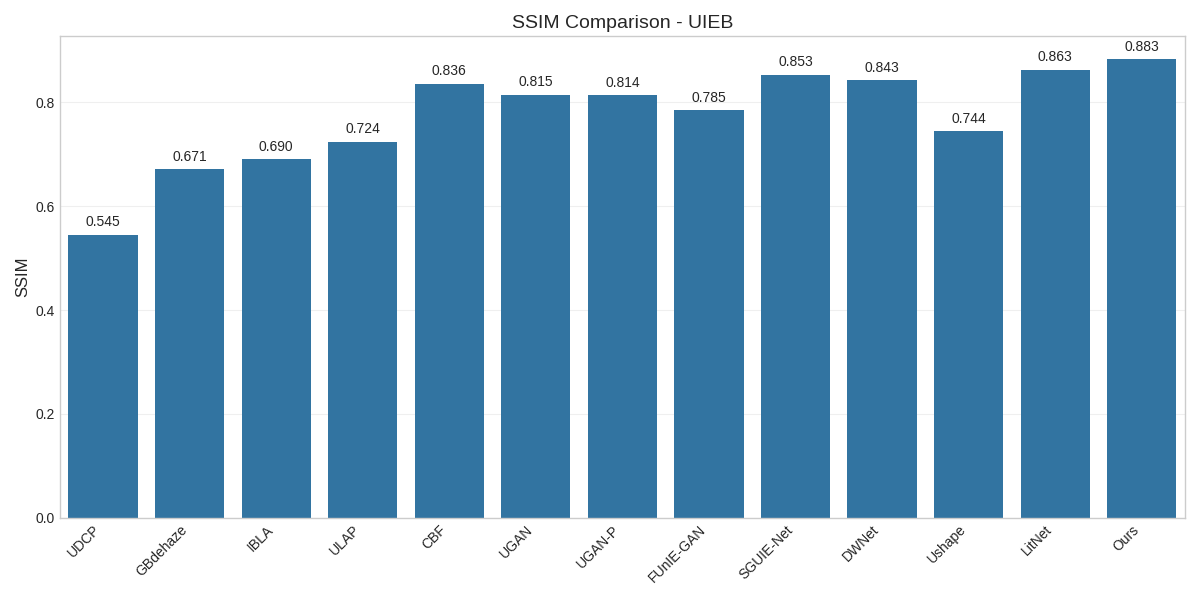}}
    \subfloat[EUVP]{\includegraphics[width=0.32\textwidth]{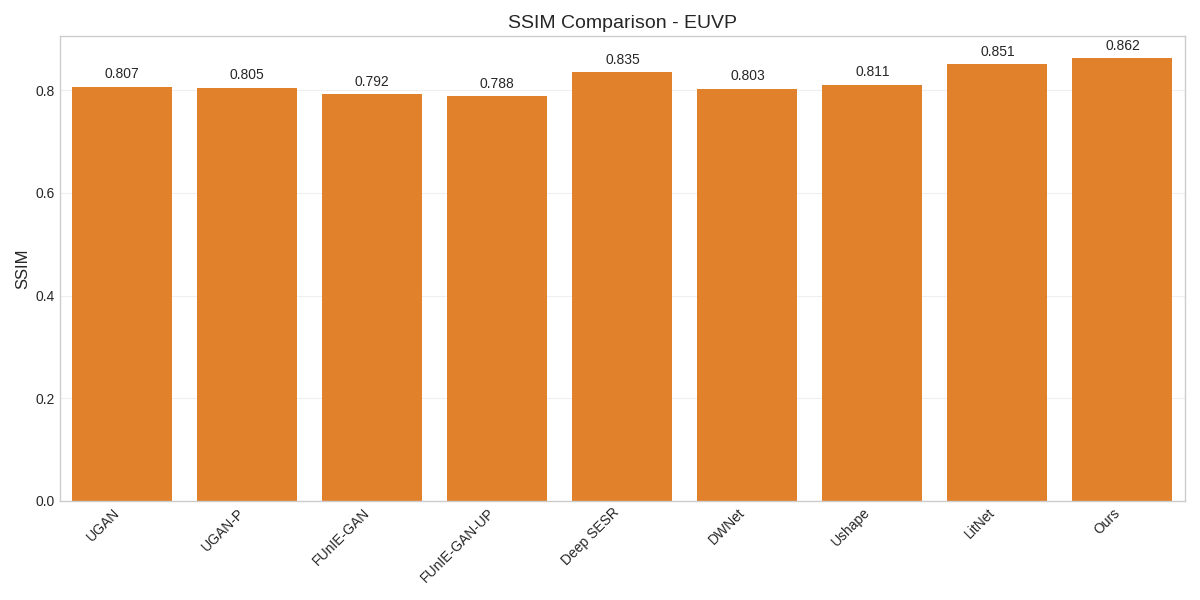}}
    \subfloat[SUIM-E]{\includegraphics[width=0.32\textwidth]{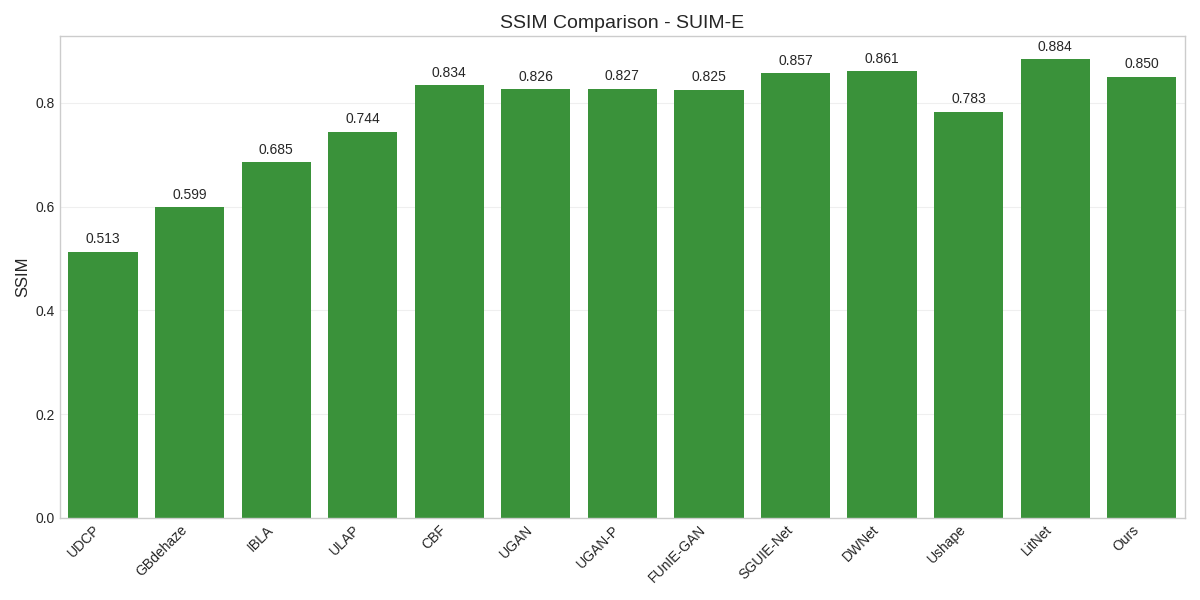}}
    \caption{Bar plots comparing SSIM values across the three datasets. FUSION consistently achieves higher SSIM values.}
    \label{fig:bar_ssim}
\end{figure*}

\begin{figure*}[!t]
    \centering
    \subfloat[UIEB]{\includegraphics[width=0.32\textwidth]{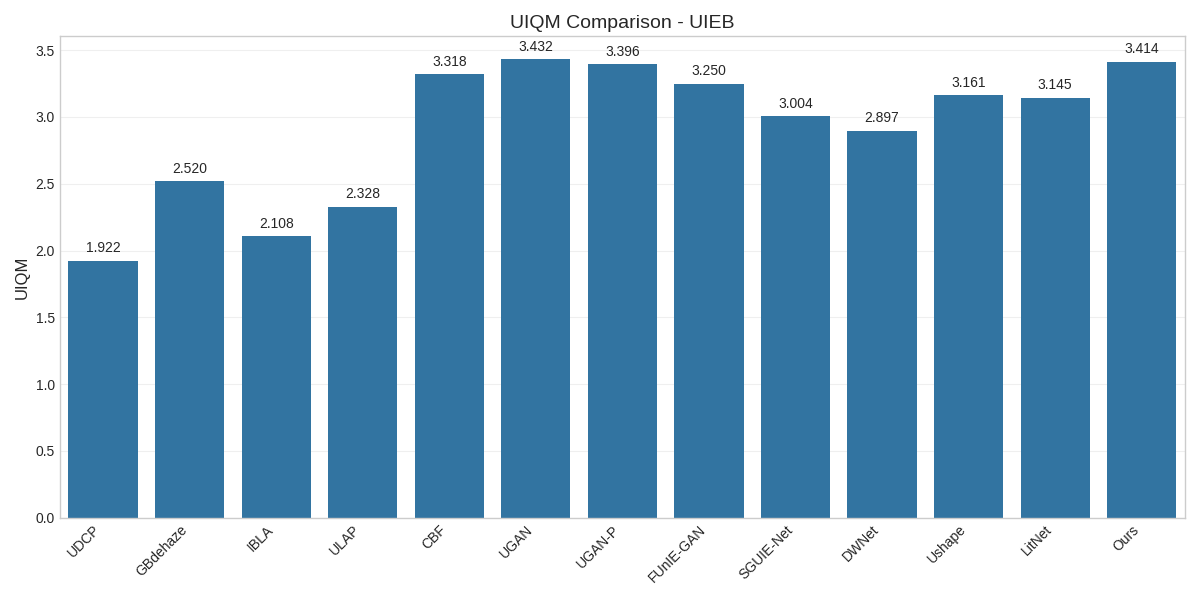}}
    \subfloat[EUVP]{\includegraphics[width=0.32\textwidth]{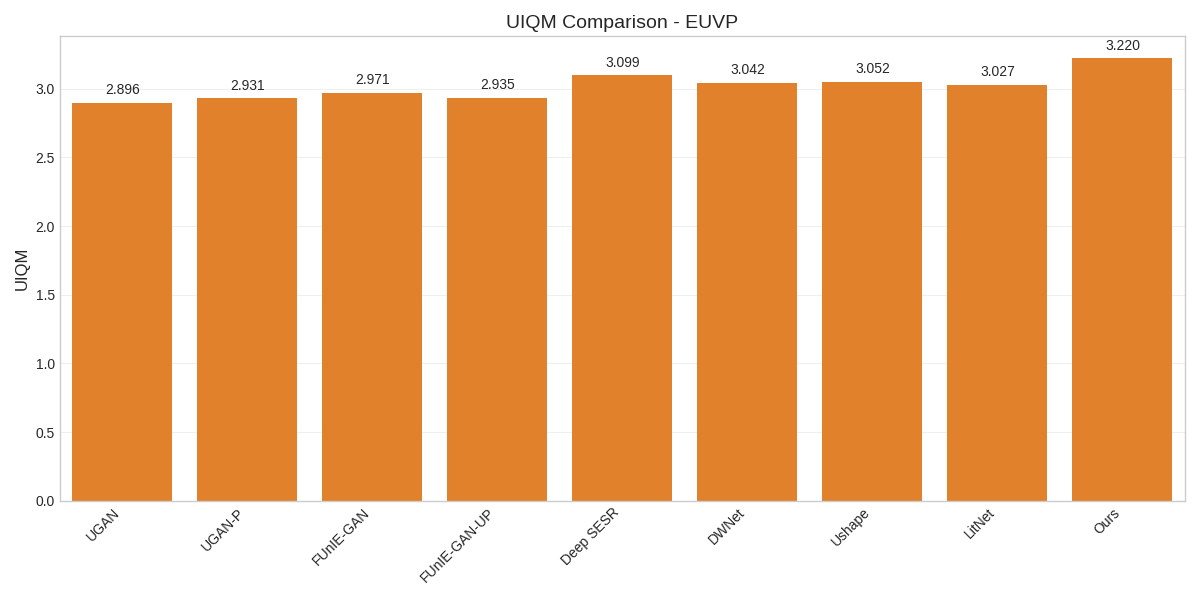}}
    \subfloat[SUIM-E]{\includegraphics[width=0.32\textwidth]{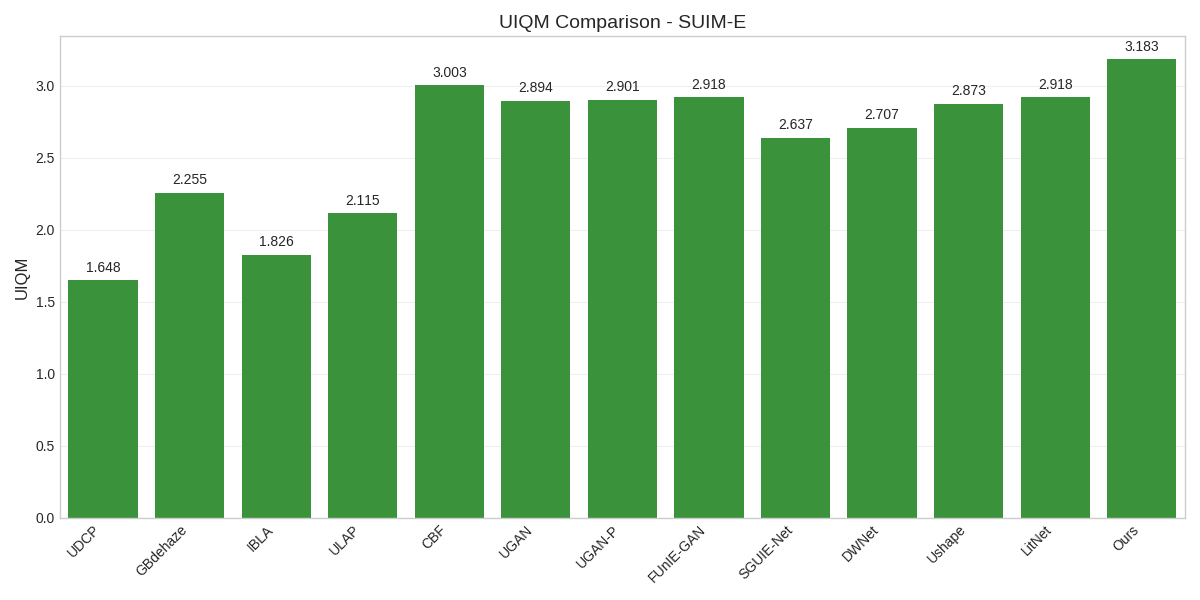}}
    \caption{Bar plots comparing UIQM scores across the UIEB, EUVP, and SUIM-E datasets. The UIQM metric reflects overall image quality improvements achieved by FUSION.}
    \label{fig:bar_uiqm}
\end{figure*}

\begin{figure*}[!t]
    \centering
    \subfloat[UIEB]{\includegraphics[width=0.32\textwidth]{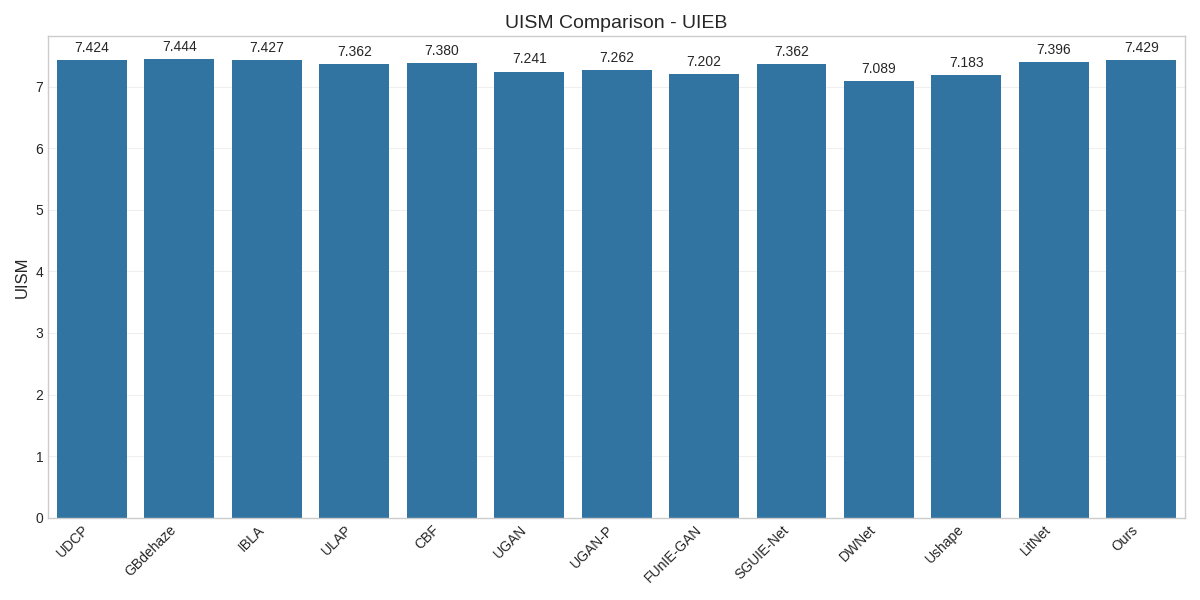}}
    \subfloat[EUVP]{\includegraphics[width=0.32\textwidth]{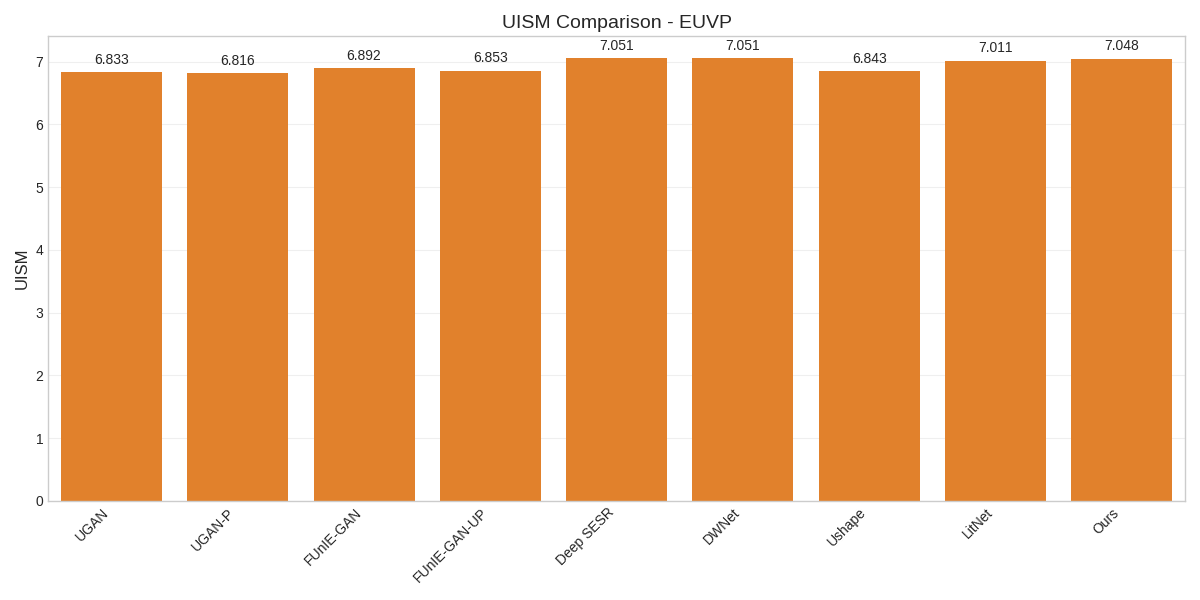}}
    \subfloat[SUIM-E]{\includegraphics[width=0.32\textwidth]{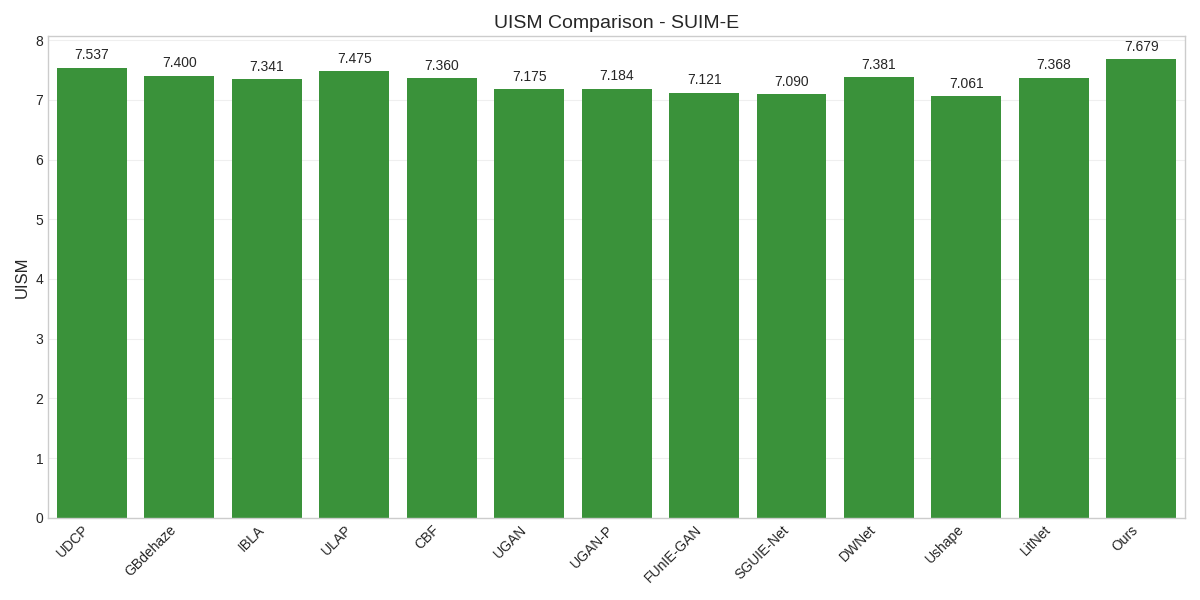}}
    \caption{Bar plots comparing UISM scores across the three datasets. Higher UISM scores for FUSION indicate improved sharpness and detail retention in the enhanced images.}
    \label{fig:bar_uism}
\end{figure*}

To provide a granular view of the performance across different metrics, we present bar plots (Figures~\ref{fig:bar_brisque}-\ref{fig:bar_uism})for each quality measure across the UIEB, EUVP, and SUIM-E datasets. These plots allow us to compare how various methods perform in terms of perceptual quality (BRISQUE and LPIPS), reconstruction fidelity (PSNR and SSIM), reconstruction (MSE), and overall image quality (UIQM and UISM).

\subsection*{Additional Efficiency Analysis}
The efficiency of underwater image enhancement models is crucial for practical deployment, particularly on autonomous underwater vehicles (AUVs) and other resource-constrained platforms. In this section, we provide a side-by-side comparison of the model parameters and computational complexity (GFLOPs) for various SOTA methods. As described in our methodology, the efficient design of FUSION is achieved by leveraging dual-domain processing and optimized fusion strategies, which are mathematically formulated in Equations (1) through (9) for the spatial and frequency domains, respectively.

Figure~\ref{fig:side_efficiency} illustrates the trade-off between the parameter count and GFLOPs. This side-by-side visualization clearly shows that FUSION achieves competitive computational efficiency, with a remarkably low parameter count (0.28M) while maintaining a GFLOPs score of 36.73G. This balance is a direct result of the adaptive attention mechanisms and efficient convolutional designs implemented within the network.

% \subsection*{PSNR vs. GFLOPs Analysis}
% To further explore the performance-efficiency trade-off, Figure~\ref{fig:psnr_flops_sup} presents a bubble chart depicting the relationship between average PSNR and GFLOPs for different UIE models. In this visualization, each bubble represents a model: its position along the GFLOPs axis indicates the computational cost, while the corresponding PSNR value reflects reconstruction quality. Additionally, the bubble size can be interpreted as an indicator of model complexity (parameter count), offering a comprehensive view of how each method balances performance and efficiency.

% This bubble chart underscores the effectiveness of FUSION in achieving high-quality image enhancement while keeping computational demands moderate. The favorable positioning of FUSION—high PSNR with relatively low GFLOPs—demonstrates the practical benefits of our dual-domain design and efficient feature fusion mechanisms.

% \begin{figure}[!t]
%     \centering
%     \includegraphics[width=\columnwidth]{supplementary/psnr_flops_sup.png}
%     \caption{Bubble chart illustrating the trade-off between average PSNR and GFLOPs for various UIE models. FUSION's high PSNR and competitive GFLOPs illustrate its excellent performance-efficiency balance.}
%     \label{fig:psnr_flops_sup}
% \end{figure}

\clearpage

\end{document}